\documentclass[10pt,twocolumn,letterpaper]{article}

\usepackage[pagenumbers]{cvpr} % To force page numbers, e.g. for an arXiv version

\usepackage[dvipsnames]{xcolor}

\definecolor{cvprblue}{rgb}{0.21,0.49,0.74}
\usepackage[pagebackref,breaklinks,colorlinks,citecolor=cvprblue]{hyperref}
\usepackage{multirow}
\def\confName{CVPR}
\def\confYear{2024}

\title{RadSimReal: Bridging the Gap Between Synthetic and Real Data in Radar Object Detection With Simulation}

\author{Oded Bialer\footnotemark[1] \space  and Yuval Haitman\thanks{Both authors contributed equally to this work.\\ Both authors are with General Motors, Yuval Haitman is also with the School of Electrical and Computer Engineering in Ben Gurion University of the Negev.} \\
General Motors, 
Technical Center Israel\\
{\tt\small oded.bialer8@gmail.com, \tt\small haitman@post.bgu.ac.il}}

\begin{document}
\maketitle
\begin{abstract}
Object detection in radar imagery with neural networks shows great potential for improving autonomous driving. However, obtaining annotated datasets from real radar images, crucial for training these networks, is challenging, especially in scenarios with long-range detection and adverse weather and lighting conditions where radar performance excels. To address this challenge, we present \textit{RadSimReal}, an innovative physical radar simulation capable of generating synthetic radar images with accompanying annotations for various radar types and environmental conditions, all without the need for real data collection. Remarkably, our findings demonstrate that training object detection models on \textit{RadSimReal} data and subsequently evaluating them on real-world data produce performance levels comparable to models trained and tested on real data from the same dataset, and even achieves better performance when testing across different real datasets. \textit{RadSimReal} offers advantages over other physical radar simulations that it does not necessitate knowledge of the radar design details, which are often not disclosed by radar suppliers, and has faster run-time. This innovative tool has the potential to advance the development of computer vision algorithms for radar-based autonomous driving applications. Our GitHub: \href{https://yuvalhg.github.io/RadSimReal}{https://yuvalhg.github.io/RadSimReal}.
\end{abstract}    
\section{Introduction}\label{sec:intro}
Automotive radar plays an important role in autonomous driving systems, offering long-range object detection capabilities and robustness against challenging weather and lighting conditions. The radar emits radio frequency (RF) signals and, through the processing of reflected echoes from the surrounding environment, creates a radar reflection intensity image \cite{bilik2016automotive}. The image contains reflection intensities corresponding to range and angle coordinates, providing a  visual representation of the scene. Afterwards, computer vision algorithms are employed to identify objects within this visual image.

\begin{figure}
  \centering
   \includegraphics[width=1\linewidth]{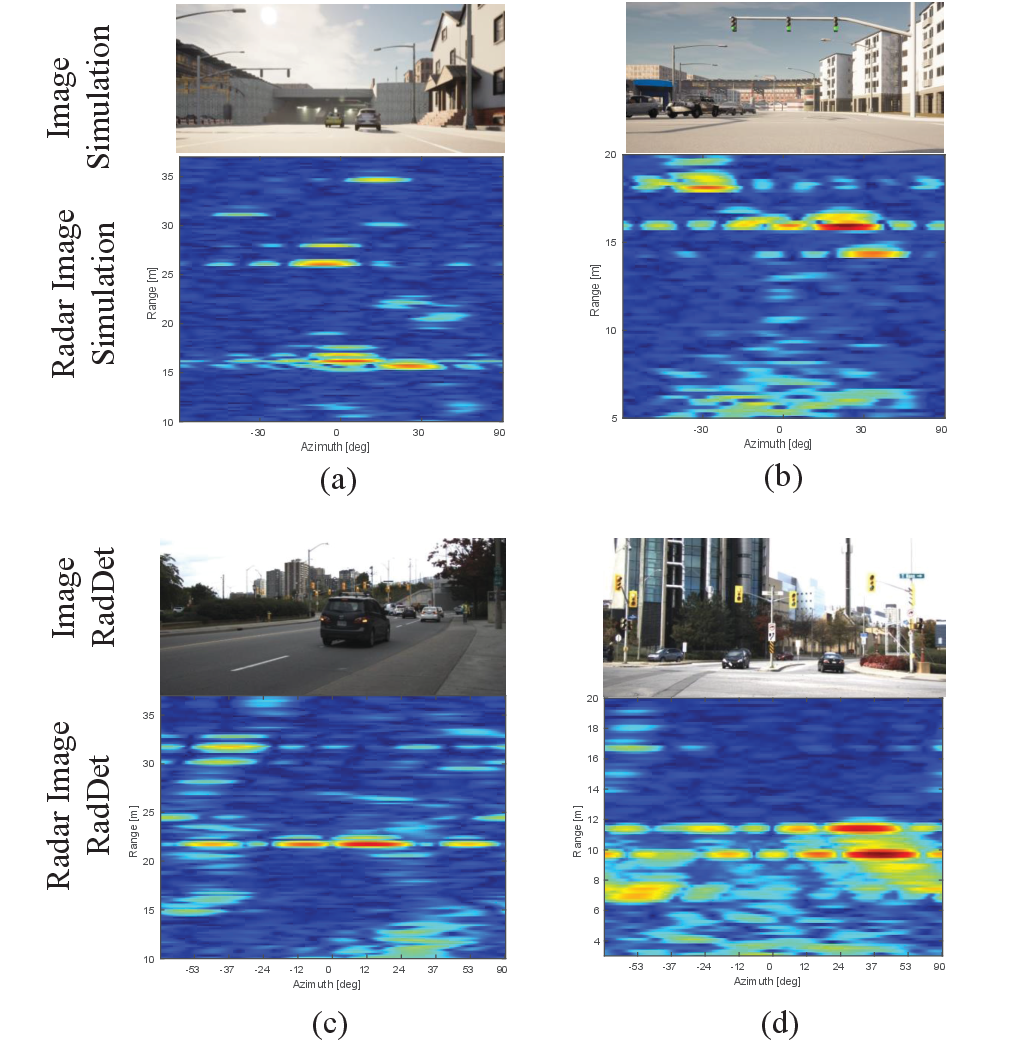}   
   \vspace{-21pt}
   \caption{Comparison between synthetic and real radar images from four different scenarios. Each scenario shows the camera image and the corresponding radar image. (a) and (b) simulation scenarios. (c) and (d) real scenarios.}
\label{fig:sim_validation_diff_scenarios}
\vspace{-10pt}
\end{figure}

Numerous Deep Neural Network (DNN) methods have emerged for detecting objects in radar images \cite{kim2020yolo,redmon2016you,dong2020probabilistic,zhang2021raddet,zhang2020object,meyer2021graph}. These techniques involve training the DNN using annotated real data. Several datasets containing real annotated radar images have been introduced \cite{zhang2021raddet,sheeny2021radiate,ouaknine2021carrada,wang2021rodnet,rebut2022raw,paek2022k}. These datasets vary in terms of the radar type and environmental conditions. However, the primary challenge with object detection DNNs trained on real data lies in the considerable effort required to collect and annotate the data. This challenge is particularly hard in the case of radar since it is used to detect objects at long range, adverse weather and lightening conditions in which annotations are difficult to obtain. 

In an effort to address the challenge posed by data annotations, an alternative approach that generates training data through generative methods has been proposed. Several studies have explored the training of a Generative Adversarial Network (GAN) using unlabeled real radar data to produce synthetic radar data that closely mimics actual real radar data \cite{weston2021there,fidelis2023generation,de2020generating,wheeler2017deep}. These studies have demonstrated that when training a detection DNN with synthetic data generated by the GAN and testing on real data, the performance gap compared to training with real data is small.

Despite the advantage of not requiring data annotation, generative data generation still presents the hurdle of collecting a large volume of unlabeled real data. This poses a significant limitation in system development, as it necessitates collecting a substantial domain-specific dataset for each unique radar sensor, distinct sensor mounting conditions, and environmental conditions in order to effectively train the GAN for generating radar images that match the specific distribution of the data. This problem is resolved when using physical radar simulation instead of generative radar simulation.

In physical simulation, synthetic radar images are created through the physical modeling of the environment and the radar sensor \cite{schoffmann2021virtual,arnold2022maxray}. Consequently, for each distinct radar sensor, mounting setup, scenario distribution, and environmental conditions a simulated dataset can be generated without the necessity of collecting any real radar data. Physical radar simulations have been extensively explored in the radar domain \cite{schoffmann2021virtual,arnold2022maxray,schussler2021realistic,thieling2020scalable,hirsenkorn2017ray,holder2019fourier}. Their process involves several steps. It begins with the creation of 3D automotive scenarios, followed by the calculation of radar reflections achieved through signal ray tracing from the radar to objects and back. Subsequently, the radar's received signal is generated based on these reflections and the specific radar hardware configuration. Finally, the radar image is produced by applying radar-specific signal processing algorithms to the received signal. 

The specific hardware and signal processing design of a radar significantly influence the output radar image. Thus, the domain shift from one radar type to another is significant, possibly more pronounced than in other sensors like cameras. This underscores a major limitation of current physical radar simulations, as they demand a comprehensive understanding of the radar's hardware parameters and signal processing algorithms to produce a radar image that accurately replicates the real-world image. These details are not always disclosed by radar suppliers, and even when available, their implementation within the simulation necessitates radar expertise. Another issue with physical radar simulation is its high computational demand, resulting in long processing times when generating large datasets.

The similarity between real radar images and synthetic images generated by physical radar simulation, when the hardware and signal processing details of the radar are available, has been evaluated both qualitatively \cite{schussler2021realistic,hirsenkorn2017ray,holder2019fourier} and quantitatively.
The quantitative evaluation involves calculating correlation coefficients between synthetic and real images \cite{schoffmann2021virtual}, and measuring distances between corresponding reflection points or objects in synthetic and real images \cite{ngo2021multi}. However, there has been a lack of evaluation concerning the performance gap between an object detection DNN trained with physical radar data and tested on real data, in comparison to a DNN solely trained and tested on real data. 

In this paper, we present \textit{RadSimReal} an innovative physical radar simulation method that does not require prior knowledge of the radar hardware specifications or its signal processing algorithms. Consequently, it can be applied to a wide range of radar sensors without necessitating expertise in radar-specific details. Additionally, this novel simulation approach offers considerably faster processing time in comparison to conventional physical radar simulations. 
Fig.~\ref{fig:sim_validation_diff_scenarios} illustrates the similarity between radar images generated by \textit{RadSimReal} and real radar images, which raises the question of whether synthetic images can replace real ones for training object detection neural networks. To answer this, we conduct a novel analysis comparing the performance of DNN object detection models trained with \textit{RadSimReal} data and tested on real data versus models trained and tested exclusively with real data.

Our findings reveal that object detection DNNs trained on \textit{RadSimReal} data and tested on real data exhibit performance levels comparable to those trained on real data when both the training and testing datasets are sourced from the same real dataset, and outperform them when the training and testing are from different real datasets. \textit{RadSimReal} is a powerful tool for efficiently generating extensive annotated training data with  flexibility in radar sensor type, mounting configurations and environmental conditions, all without the overhead effort of collecting real data for each specific setting.

The main contributions of the paper are as follows:
\begin{enumerate}
\item A pioneer analysis revealing that object detection DNNs, trained with physical radar simulation data and tested on real data, perform comparably to models trained on real data when the train and test sets of the real data are from the same dataset, and outperforms them when they are from different datasets. 
This highlights the benefits of physical simulation, efficiently generating training data with annotations for diverse radar configurations without the challenges of collecting real data, overcoming a main limitation in generative approaches.

\item An innovative physical radar simulation technique that offers an important advantage over other reference physical simulation methods by not necessitating in-depth knowledge of specific radar implementation details, which are often undisclosed, nor radar expertise, and also has a significantly faster run-time.
\item A simulation tool offering efficient data generation with ground truth information that conveniently supports various radar types. This tool will contribute to advancing radar-based computer vision research.
\end{enumerate}
\section{Related Work}\label{sec:related_work}
\subsection{Radar Object Detection}
Various studies in the literature extensively investigate DNN methods for object detection in radar images. Kim et al. \cite{kim2020yolo} applied YOLO \cite{redmon2016you} to radar images, surpassing the performance of conventional radar detection methods. Xu et al. \cite{dong2020probabilistic} used a ResNet-18 encoder with a CNN decoder to estimate 3D bounding box properties. RADDet \cite{zhang2021raddet} integrated residual blocks and YOLO detection heads \cite{bochkovskiy2020yolov4}, while Zhang et al. \cite{zhang2020object} employed a CNN-based version of U-Net for radar object detection. Meyer et al. \cite{meyer2021graph} utilized graph convolution networks for radar object detection. A two-stage object detection approach was introduced in \cite{haitman2024boostrad}. An alternative approach for object detection involves using radar point clouds \cite{danzer20192d, schumann2018semantic, scheiner2019radar, feng2019point, kraus2020using, dreher2020radar}, which are generated from the reflection points detected in radar images using the Constant False Alarm Rate (CFAR) algorithm \cite{rohling1983radar}. CFAR introduces significant information loss during subsequent DNN processing \cite{kim2020yolo}, leading to a considerable degradation in object detection performance when compared to using radar images. Therefore, this paper focuses on DNN processing of radar images.

\subsection{Radar Datasets}
Numerous publicly accessible automotive radar datasets offer real radar images along with object annotations. The RADIATE \cite{sheeny2021radiate} and Oxford RobotCar \cite{barnes2020oxford} datasets use a 360° mechanical scanning antenna, differing from conventional radar images generated by antennas arrays.
The CRUW dataset \cite{wang2021rodnet2} provides radar images with a limited range of up to 25 meters, and the annotations consist of object center points without bounding box information. RADIal \cite{rebut2022raw} and K-Radar \cite{paek2022k} provide high-resolution radar reflection intensity images, but details about their radar hardware are undisclosed, making them unsuitable for conventional physical radar simulation. 
The RADDet \cite{zhang2021raddet} dataset features radar images from 15 diverse automotive scenarios captured using a Texas Instruments (TI) automotive radar prototype \cite{TI_PROTOTYPE_1,TI_PROTOTYPE_2}. The CARRADA dataset \cite{ouaknine2021carrada} comprises radar images from 30 controlled scenarios, utilizing the same TI radar as RADDet, allowing for cross-dataset performance evaluation. Furthermore, the availability of hardware specifications for the TI prototype radar in these datasets facilitates the generation of synthetic data through traditional physical radar simulation.
\vspace{-4.5pt}

\subsection{Generative Radar Data Generation}
Weston et al. \cite{weston2021there} introduced a GAN method that generates synthetic radar images conditioned on a 3D representation from CARLA simulation \cite{dosovitskiy2017carla}, which was  trained on a relatively large dataset comprising 222,420 images from the Oxford RobotCar \cite{barnes2020oxford}.
Their study showed that a segmentation model, trained on synthetic data and tested on real data, achieves performance similar to a model trained and tested exclusively on real data.
L2R GAN \cite{wang2020l2r} used the same extensive dataset to train a GAN for converting LIDAR point clouds into radar images. Synthetic image quality was evaluated against real images using PSNR and SSIM metrics, along with qualitative assessment through a human subjective study. Oliveira and Bekooij \cite{de2020generating} employed a GAN to generate synthetic radar images conditioned on bounding box layouts. They showed a small performance gap between an object detection DNN trained on synthetic data and tested on real data versus one trained solely on real data. Fidelis \cite{fidelis2023generation} used a GAN to generate radar received signals and derived radar images through signal processing. Their study showed a close distribution resemblance between synthetic and real images, assessed by the FID metric \cite{heusel2017gans}.
Wheeler et al. \cite{wheeler2017deep} used a Variational Autoencoder conditioned on an object list and raster grid of the ground roadways. They validated synthetic image similarity to real images using K-L divergence for clutter and average squared deviation for objects. While generative methods bridge the synthetic-real data gap, they require extensive training data for each radar type, mounting configuration, and environmental condition, posing a significant overhead.

\subsection{Physical Radar Simulation Data Generation}
In physical simulation, synthetic radar images are generated by simulating the environment, the radar sensor, and its installation on the vehicle. Numerous studies have proposed methods for physically modeling both the environment and radar systems.
MaxRay \cite{arnold2022maxray} created realistic scenarios in Blender \cite{hess2013blender}, simulated reflections through ray tracing with RF propagation properties, and tested radar detection and clutter removal on synthetic data of an OFDM radar simulation.
ViRa \cite{schoffmann2021virtual} used the Unity game engine \cite{buyuksalih20173d} to generate environments and simulated an FMCW radar. They demonstrated the similarity of their simulated radar data to real-world radar measurements in a laboratory setting using correlation metrics. Thieling et al. \cite{thieling2020scalable} proposed a radar simulation with environmental influences like rain. Their study assessed the simulation's performance in qualitative terms. In another approach \cite{schussler2021realistic}, a ray tracing technique for a MIMO radar simulation was presented, with the realism of the generated radar images evaluated through qualitative comparisons to real radar images for the same scenarios. Several other ray tracing radar simulations for realistic automotive scenarios have also been published \cite{hirsenkorn2017ray, dudek2010millimeter, holder2019fourier}, which verified the simulations' accuracy by comparing them to real measurements. 

All these radar simulation studies haven't explored the use of DNN detection methods with physical radar simulation data. The performance gap between training object detection DNNs with radar simulation synthetic data versus real data and testing on real data remains unexplored. Furthermore, implementing these simulations requires detailed knowledge of radar hardware design and signal processing algorithms, factors often undisclosed by radar suppliers.
\section{Proposed Physical Simulation Method}\label{sec:sim_metthod}
\begin{figure*}[t]
  \centering
   \includegraphics[width=0.90\linewidth]{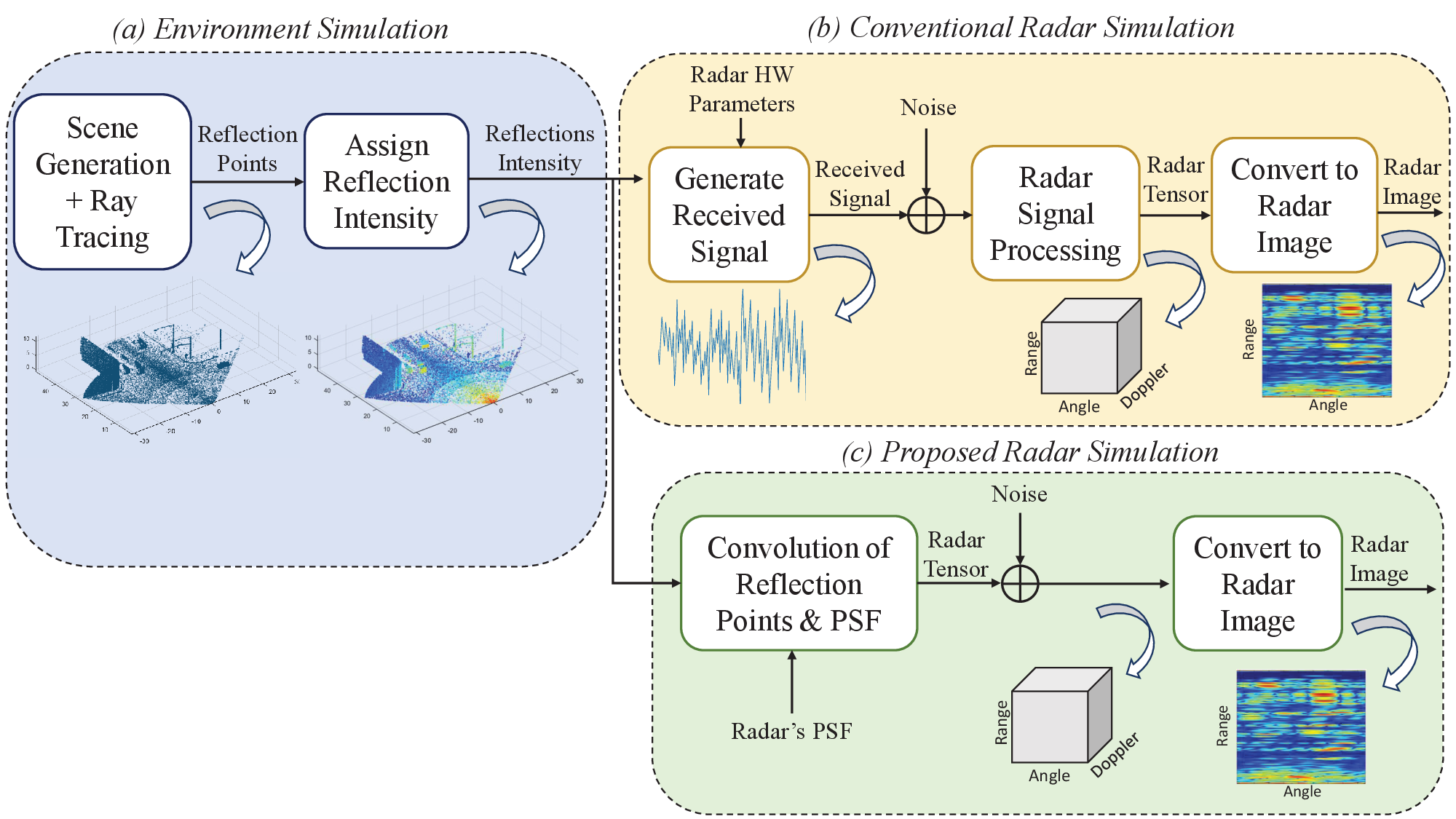}
   \caption{Block diagram illustrating the processing steps for conventional simulation (a)+(b) and \textit{RadSimReal} (a)+(c). (a) Simulates the environment to generate reflection points with RF reflectivity of an automotive scene, while (b) and (c) represent the conventional approach and \textit{RadSimReal}'s approach, respectively, for transforming the reflection points into a radar image.}   
   \label{fig:sim_block_diagram}    
   \vspace{-5pt}
\end{figure*}
Fig.~\ref{fig:sim_block_diagram} presents a block diagram comparing our proposed physical simulation approach, \textit{RadSimReal}, with the conventional physical simulation method. The diagram consists of three primary components. \textit{RadSimReal} encompasses Fig.~\ref{fig:sim_block_diagram}(a) and (c), whereas the conventional simulation method involves Fig.~\ref{fig:sim_block_diagram}(a) and (b). Fig.~\ref{fig:sim_block_diagram}(a) pertains to the environmental simulation, a shared element in both approaches, independent of the specific radar sensor.
In this part, a 3D scene is generated using a graphics engine; in our implementation, we employed CARLA \cite{dosovitskiy2017carla}. Then, dense reflection points from objects within the scene are acquired by ray-tracing of RF propagation paths, starting from the radar to objects in the environment and then returning to the radar \cite{schussler2021realistic}. Subsequently, the RF reflectivity of these points is determined using physical formulas \cite{bassem2004matlab}, accounting for surface material, orientation, and distance from the radar. Following the environmental simulation, the subsequent step involves the radar simulation part, responsible for translating the intensities of reflection points into the radar's output image. Fig.~\ref{fig:sim_block_diagram}(c) illustrates the radar simulation block diagram of \textit{RadSimReal}, which  deviates from the conventional radar simulation method shown in Fig.~\ref{fig:sim_block_diagram}(b).

In the conventional radar simulation presented in Fig.~\ref{fig:sim_block_diagram}(b), the radar's received signal, stemming from all reflection points within the scene, is acquired based on the specific hardware design details of the radar. Additionally, noise is introduced into the received signal. For detailed information regarding this part, we refer readers to the supplementary material. Following this, radar signal processing algorithms are applied to the received signal, yielding the radar 3D tensor. This tensor encapsulates the radar reflection intensity across range, azimuth angle, and Doppler cells. Notably, we consider a radar with only azimuth angle and without elevation, although the simulation is not limited to such radar and could be expanded to include elevation. In the final step of Fig.~\ref{fig:sim_block_diagram}(b), the 3D tensor is transformed into an image format with dimensions of range and azimuth angle by selecting the maximum value along the Doppler dimension. It's worth mentioning that the simulation provides flexibility in outputting the entire radar tensor or converting it into an image using various techniques, depending on the implementation of the object detection DNN. 

The conventional radar simulation method depicted in Fig.~\ref{fig:sim_block_diagram}(b), necessitates an in-depth comprehension of the radar hardware design and the associated software processing to faithfully mimic a specific radar. This entails a comprehensive understanding of specific details such as the transmit signal waveform, antenna array configuration, sampling rate, beamforming algorithms, and more. However, this presents a major challenge, as these details are often proprietary and not publicly disclosed by radar manufacturers. This limitation is resolved in Fig.~\ref{fig:sim_block_diagram}(c) by \textit{RadSimReal}, as elaborated in the following.

\begin{figure}[h]
  \centering
   \includegraphics[width=0.90\linewidth]{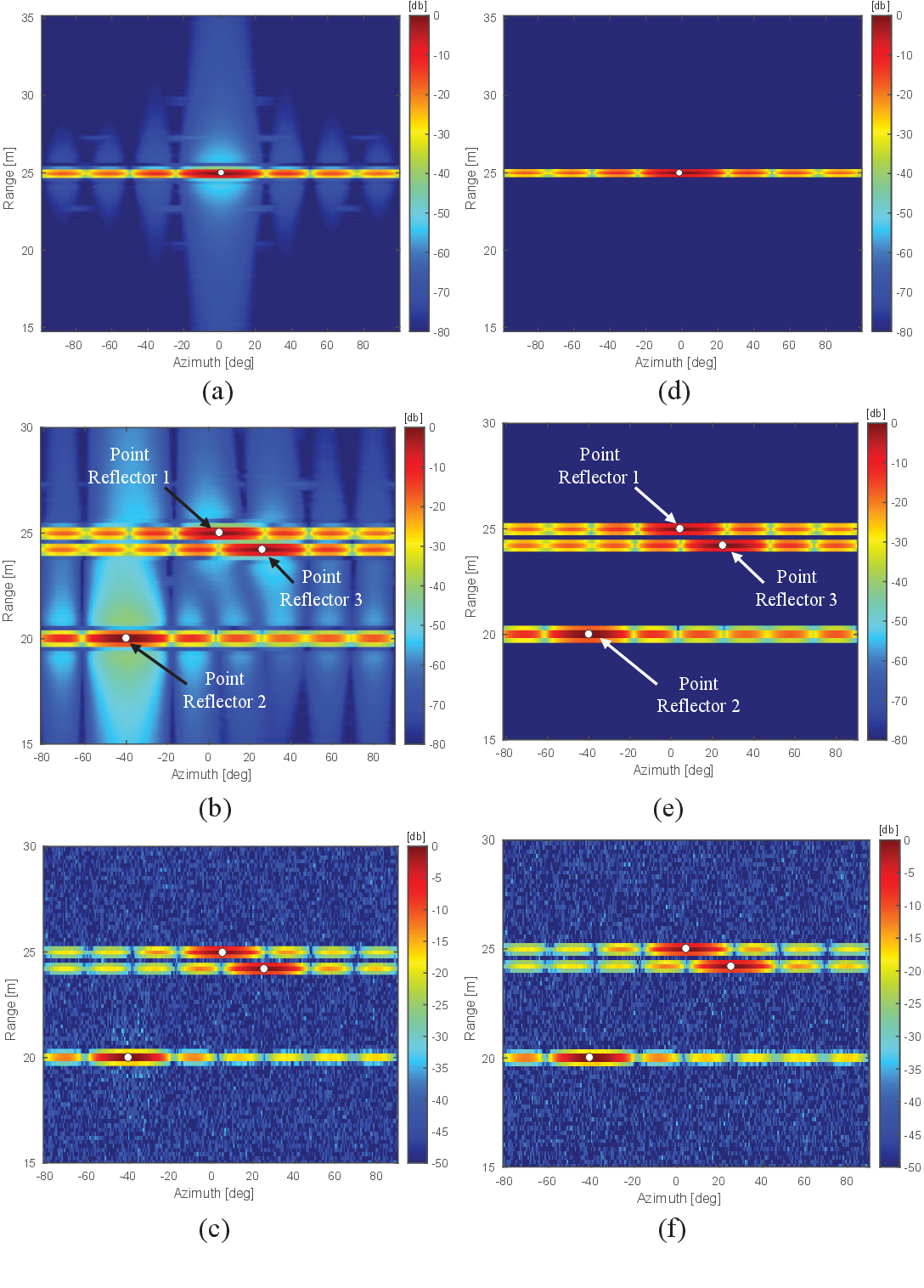}
   \vspace{-10pt}
   \caption{Radar image generated with conventional simulation vs. \textit{RadSimReal}. (a) Radar's PSF 2D slice in range and angle dimensions. (b) Radar image without noise generated by conventional simulation for a scenario with 3 reflection points. (c) Radar image of (b) with noise. (d) Truncated PSF with $99\%$ of its energy. (e) Radar image obtained for the same scenario as in (b) by \textit{RadSimReal} without noise, (f) The radar image of (e) with noise.}  
   \label{fig:sim_convolution_illustration}
   \vspace{-10pt}
\end{figure}

\textit{RadSimReal} relies on the radar's point spread function (PSF). Every reflection point is apparent in the radar tensor by a multi-dimensional PSF, spanning dimensions of range, angle, and Doppler, and centered on the coordinates of the reflection point. Fig.~\ref{fig:sim_convolution_illustration}(a) shows an example of a 2D slice of a radar's PSF in range and angle centered at a range of $25m$ and an azimuth angle of $0^{\circ}$. The PSF exhibits a wide spread in the angular domain but a narrow spread in both range and Doppler, although the Doppler aspect is not depicted in the figure. This characteristic arises from the radar's coarse angular resolution and high range and Doppler resolution. It is essential to note that the shape of the PSF depends upon the specific radar hardware design and signal processing algorithms \cite{skolnik1980introduction}. Fortunately, it can be acquired from a radar tensor of a narrow object, such as a pole or a dedicated corner reflector \cite{knott2004radar}. Importantly, obtaining the PSF does not require knowledge of any details about the radar design, thereby circumventing a key limitation of conventional radar simulation. The straightforward procedure for obtaining the radar's PSF from a radar tensor is detailed in the supplementary material.

In Fig.~\ref{fig:sim_convolution_illustration}(b), we present a radar image generated through the conventional simulation outlined in Fig.~\ref{fig:sim_block_diagram}(b). This image illustrates a scenario with three reflection points and is free from noise. In Fig.~\ref{fig:sim_convolution_illustration}(c), we depict the outcome of the conventional simulation for the same scenario, this time with the addition of noise. The white points in the figures denote the locations of these reflection points. Notably, the image displays a superposition of the radar's PSF centered on each point. Therefore, the same image can be obtained by convolution between the reflection points and the PSF. As a result, \textit{RadSimReal} generates the radar tensor by 3D convolution of the reflection points with the PSF, followed by the addition of noise, as presented in Fig.~\ref{fig:sim_block_diagram}(c). The tensor is then converted to an image similarly as in Fig.~\ref{fig:sim_block_diagram}(b).

Despite the differences in the simulation implementations in Fig.~\ref{fig:sim_block_diagram}(c) and Fig.~\ref{fig:sim_block_diagram}(b), they lead to the generation of an identical radar image when the full PSF is applied.
For a detailed mathematical explanation of this equivalence, please refer to the supplementary material. Notably, the energy of the PSF diminishes rapidly from its central point in both the range and Doppler domains. To reduce the computational complexity of the simulation, we truncate the PSF to encompass up to $99\%$ of its energy. This truncation leads to a substantial reduction in the PSF size by a factor exceeding 1000, thereby significantly enhancing the run-time of the convolution operation in the simulation. In Fig.~\ref{fig:sim_convolution_illustration}(d), the truncated PSF is displayed, containing $99\%$ of the energy of the original PSF from Fig.~\ref{fig:sim_convolution_illustration}(a). The radar image, obtained through convolution of the truncated PSF with the same three reflection points as depicted in Fig.~\ref{fig:sim_convolution_illustration}(b), is showcased in Fig.~\ref{fig:sim_convolution_illustration}(e) without noise and in Fig.~\ref{fig:sim_convolution_illustration}(f) with the addition of noise. The truncated portion of the PSF has very low intensity value (about 80 dB lower than PSF peak), which is significantly lower than the noise level. Consequently, the difference between the radar image from the conventional simulation (Fig.~\ref{fig:sim_convolution_illustration}(c)) and our proposed simulation utilizing the truncated PSF (Fig.~\ref{fig:sim_convolution_illustration}(f)) is practically indistinguishable.

While \textit{RadSimReal} (Fig.~\ref{fig:sim_block_diagram}(a)+(c)) produces radar images similar to the conventional simulation (Fig.~\ref{fig:sim_block_diagram}(a)+(b)), it possesses two significant advantages over the conventional simulation. 
Firstly, it eliminates the necessity of possessing in-depth radar design information by relying on a simple radar measurement of its PSF and noise variance, as detailed in the supplementary material. Secondly, it exhibits a significantly faster run-time, as evaluated and demonstrated in Section \ref{sec:computatoin_analysis}.

\subsection{Simulation Fidelity Evaluation}\label{sec:sim_eval}
In this section, we assess the resemblance between real radar images and synthetic images generated by \textit{RadSimReal}. 
Our evaluation initiates with a comparison between synthetic and real radar images generated from the same scenario. Creating a simulation scene that faithfully replicates a real-world scenario for which an actual radar measurement was taken poses a considerable challenge. To overcome this challenge, we substituted the reflection points derived from the CARLA simulation engine in Fig.~\ref{fig:sim_block_diagram}(a) with points obtained from a high-resolution LIDAR sensor. This LIDAR was positioned in close proximity to the radar and captured measurements from the same scene at the same time as the real radar measurement. For the assignment of the RF reflection intensity to the LIDAR points the surface material and orientation of the LIDAR points were obtained by the following three steps: (1) Manual segmentation of LIDAR points into object types such as vehicles, poles, signs, roads, and buildings. (2) Surface material allocation of each point based on its object type. (3) Computation of the angle of the surface normal vector using a polygon mesh generated from the LIDAR points. The remaining stages of the simulation processing were as outlined in Fig.~\ref{fig:sim_block_diagram}(c).

\begin{figure}
  \centering   \includegraphics[width=1\linewidth]{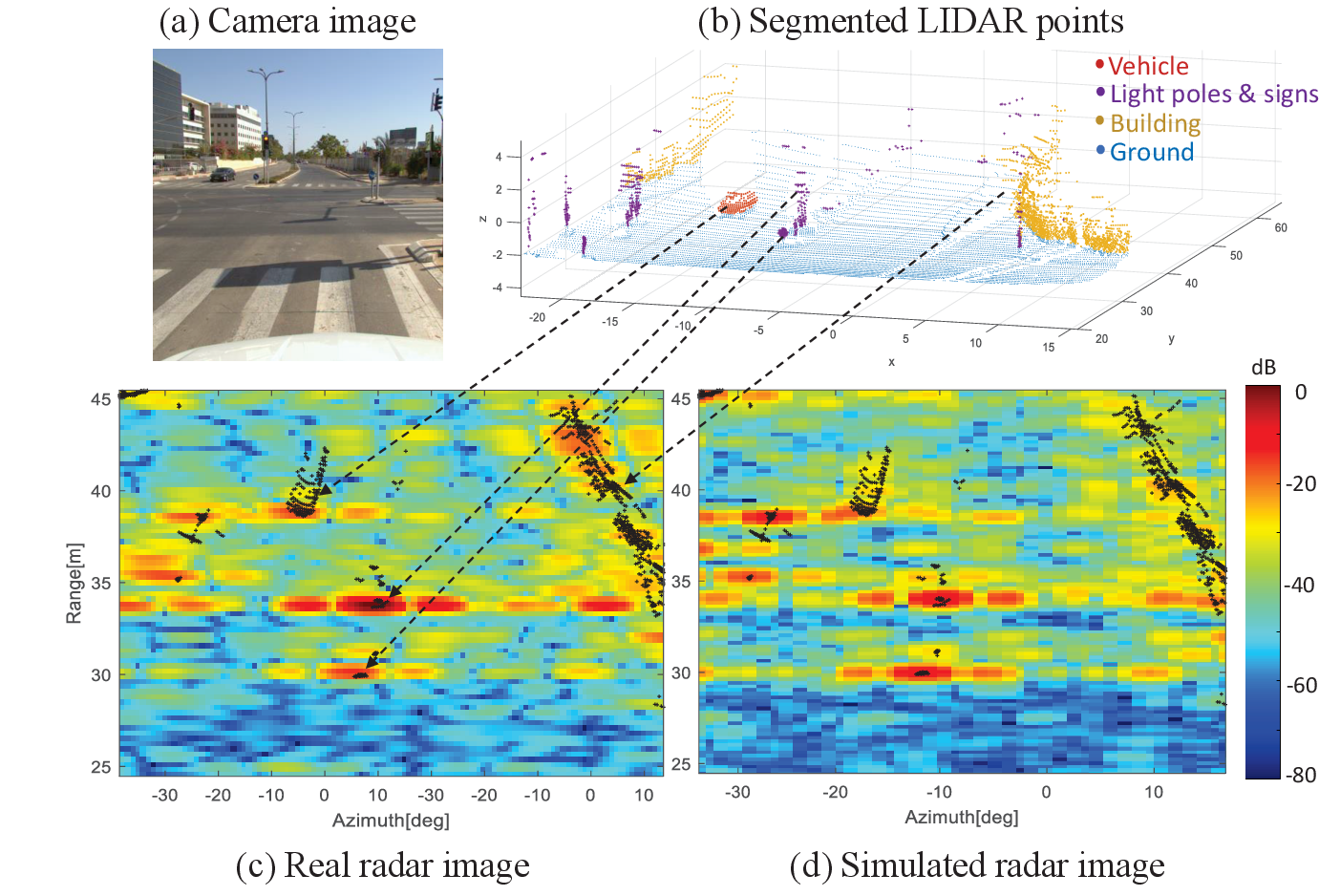}
   \vspace{-15pt}
   \caption{Comparison between \textit{RadSimReal} image and a real radar images for the same scenario. (a) Camera image of the scenario. (b) High-resolution LIDAR points segmented by object type. (c) Real radar image. (d) \textit{RadSimReal} image. The black points in (c) and (d) represent the LIDAR points.}
   \label{fig:sim_validation_same_scenario}
\end{figure}

An example of a real radar image is shown in Fig.~\ref{fig:sim_validation_same_scenario}(c) and a synthetic radar image in Fig.~\ref{fig:sim_validation_same_scenario}(d) generated from the same scene (as explained above). The prototype radar used in Fig.~\ref{fig:sim_validation_same_scenario}(c) had $3.9^{\circ}$ azimuth resolution, and $0.28m$ range resolution, and the same radar was simulated in Fig.~\ref{fig:sim_validation_same_scenario}(d). The camera image of the scene is presented in Fig.~\ref{fig:sim_validation_same_scenario}(a), and the LIDAR points segmented to different object types are shown in Fig.~\ref{fig:sim_validation_same_scenario}(b). It is observed that the synthetic and real reflection intensity images have close resembles, which shows that the simulation models well the real radar.

Next, we present a comparison between real radar images extracted from the RADDet dataset \cite{zhang2021raddet}, and synthetic images generated by simulating the same radar as in RADDet with \textit{RadSimReal}. As the RADDet dataset lacks LIDAR measurements, simulating the precise scenarios of real radar images, as depicted in Fig.~\ref{fig:sim_validation_same_scenario}, is unfeasible. Hence, we present synthetic and real radar images from different scenarios and assess their characteristic resemblance. In Fig.~\ref{fig:sim_validation_diff_scenarios}, the upper two rows illustrate synthetic radar images along with their corresponding camera images, produced by simulating the radar used in the RADDet dataset. Meanwhile, the lower two rows exhibit real radar images and their respective camera images from the RADDet dataset. Although the synthetic and real images stem from different scenarios and cannot be compared one to one, they exhibit analogous characteristics. Both real and synthetic radar images display reflections with varying intensities and have a similar spreading functions. The close resemblance between the synthetic and real images makes it hard to distinguish between them, which is another indication that \textit{RadSimReal} models well the real radar.

We also assessed the statistical similarity between the \textit{RadSimReal} data and real data using the Frechet Inception Distance (FID) \cite{heusel2017gans}. The FID score is commonly employed in the literature to quantify the statistical resemblance between synthetic and real datasets. The FID score between the RADDet training set (comprising 8196 images) and the RADDet test set (comprising 1962 images) is 6.76. On the other hand, the FID score between the RADDet training set and a synthetic dataset of 10,000 images generated by \textit{RadSimReal} is 6.54. The similarity between these two scores indicates that the statistical characteristics of the synthetic dataset produced by \textit{RadSimReal} closely resemble with those of the RADDet data.

\subsection{Computation Efficiency}\label{sec:computatoin_analysis}
The complexity ratio between conventional simulation and \textit{RadSimReal} corresponds to the ratio between the entire radar tensor volume and the PSF volume. As explained in Section \ref{sec:sim_metthod}, our simulation truncates the PSF to preserve $99\%$ of its energy, drastically reducing its volume. 
This results in a substantial complexity reduction, approximately by a factor of 1000, with \textit{RadSimReal} compared to conventional radar simulation.
Further details on the computational complexity calculation and run time measurements are available in the supplementary material.
\section{Simulation to Real Domain Gap Analysis}\label{sec:sim_validation}
In this section, we analyze the object detection performance gap between models trained with \textit{RadSimReal} data and those trained with real data, both tested on real data. For this analysis we use three different object detection methods: `U-Net', `RADDet', and `Probabilistic'. `U-Net' employs a U-Net as proposed in \cite{zhang2020object} with an additional input channel of the input image Cartesian coordinates. `RADDet' refers to the object detection network introduced in the RADDet paper \cite{zhang2021raddet}, while `Probabilistic' is the network from \cite{dong2020probabilistic}. 

\begin{table*}[h]
\caption{AP at different IOU for three object detection models trained on \textit{RadSimReal} or real data and tested on real data}
\vspace{-5pt}
\centering
\begin{tabular}{ccccccccccc}
\toprule
\multirow{2}{*}{Test Set} & \multirow{2}{*}{Train Set} &  \multicolumn{3}{c}{U-Net Model} & \multicolumn{3}{c}{Probabilistic Model} & \multicolumn{3}{c}{RADDet Model} \\
\cmidrule(lr){3-5} \cmidrule(lr){6-8} \cmidrule(lr){9-11}
& & @0.1 & @0.3 & @0.5 & @0.1 & @0.3 & @0.5 & @0.1 & @0.3 & @0.5 \\
\midrule
 \midrule
\multirow{3}{*}{RADDet} & RADDet & 84.76 & 83.01 & 55.53 &83.31 &74.80 &40.68 & 83.69 & 72.96 &47.95 \\
 & \textit{RadSimReal} & 85.63 & 82.16 & 57.64 & 82.75 & 75.83 & 52.36 & 83.48 & 73.91 & 46.63 \\
 & CARRADA & 22.33 &19.46 &14.04 & 24.58 &22.53 &16.98 &19.18 &15.37 &9.47 \\
 \midrule
\multirow{3}{*}{CARRADA} & CARRADA & 50.99 &49.00 &44.65 &50.16 &48.22 &39.51 &31.94 &26.65 &17.79 \\
 & \textit{RadSimReal} & 70.77 & 62.47 & 43.96 & 63.51 & 56.92 & 41.07 & 72.39 & 65.65 & 28.76 \\
 & RADDet & 62.40 &56.84 &30.78 &57.02 &54.23 &19.53 &69.63 &61.63 & 20.49 \\
 \midrule
\multirow{2}{*}{CRUW} & CRUW & 86.66 & 78.83 & 56.54 & 81.90 & 77.22 & 52.95 & 85.74 & 70.10 & 54.36 \\
 & \textit{RadSimReal} &  86.82 & 77.51 & 55.47 & 80.04 & 76.50 & 51.45 & 86.21 & 69.60 & 52.48 \\
\bottomrule
\end{tabular}
\vspace{-5pt}
\label{tab:ap_results_main}
\end{table*}

\begin{figure*}
  \centering
   \includegraphics[width=0.78\linewidth]{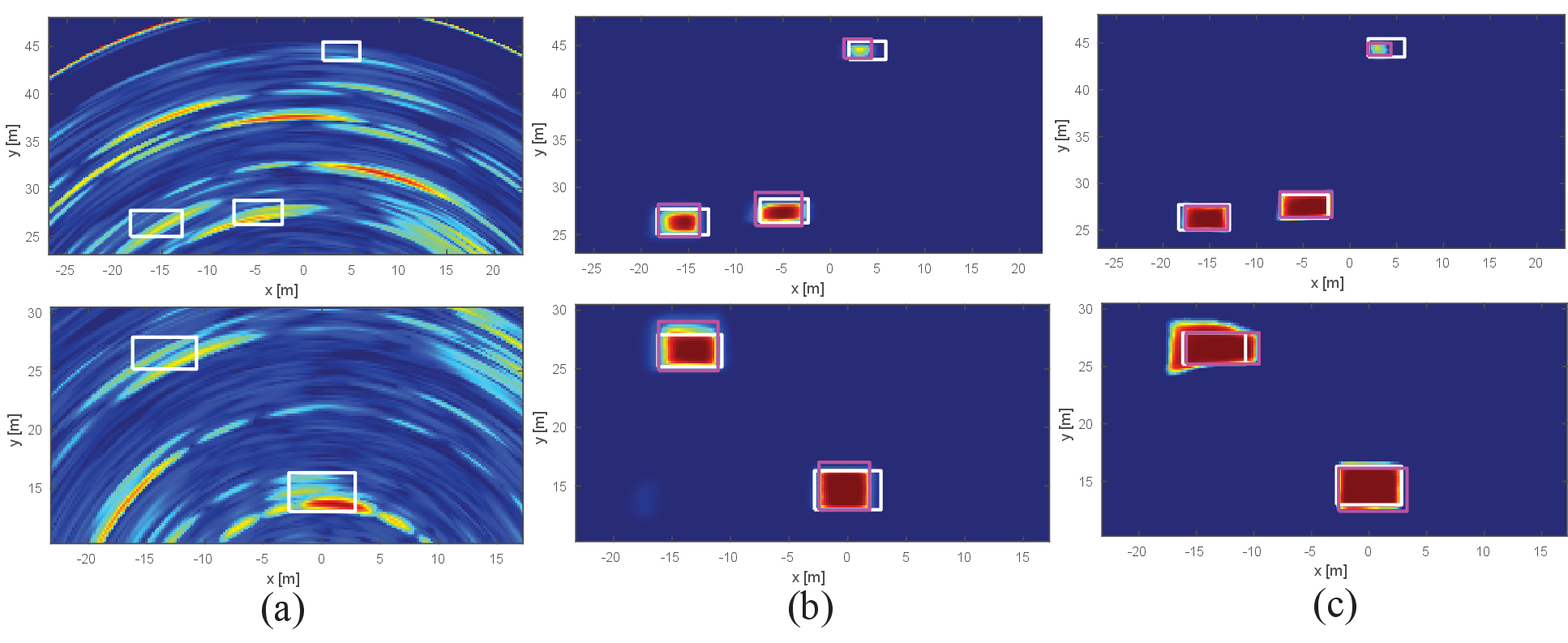}
   \vspace{-10pt}
   \caption{
   Qualitative comparison of object detection DNN trained on \textit{RadSimReal} vs. real data from the RADDet dataset. Rows correspond to different  scenarios from RADDet test set. (a) Input radar image, (b) `U-Net' model's detection score and bounding boxes trained with RADDet. (c) `U-Net' trained with \textit{RadSimReal} data. Detected and ground truth bounding boxes marked in pink and white, respectively.}
   \label{fig:qual_ex_det_score}
\end{figure*}
We assess performance using the RADDet \cite{zhang2021raddet}, CARRADA \cite{ouaknine2021carrada}, and CRUW \cite{wang2021rodnet2} datasets. These datasets feature automotive radar reflection images captured with a TI radar prototype \cite{TI_PROTOTYPE_1,TI_PROTOTYPE_2} in diverse scenes, providing extensive testing coverage. 
While RADDet and CARRADA feature 2D bounding box annotations for objects, CRUW provides annotations indicating the center points of objects, to which we've subsequently added bounding box extensions using the CFAR algorithm \cite{rohling1983radar}.
RADDet includes images from 15 densely populated automotive scenarios with favorable weather conditions, while CARRADA comprises 30 staged scenarios with varying object densities and weather conditions, including challenging conditions like snow. In both datasets the radar is mounted on a stationary platform. In the CRUW dataset the radar is mounted on a vehicle, capturing radar images from scenarios where the ego-vehicle was in motion (highway and city streets) and scenarios where it was  stationary (campus roads and parking lots).

We employed a dataset split for RADDet that ensures different scenarios in the train and test sets thereby preventing potential overfitting in the split proposed in the dataset paper \cite{zhang2021raddet}.
Our RADDet split consists of a training set with 8196 images and a test set with 1962 images. For CARRADA, the training set comprises 2208 images, and the test set includes 276 images. As for CRUW, our training set comprises 9623 images, with a test set of 2226 images. The synthetic dataset generated by \textit{RadSimReal} comprised 10000 training images, comparable in size to that of RADDet and CRUW.
The simulated scenarios involved a radar mounted on a vehicle driving in city streets, experiencing both stationary and moving phases.

We conducted performance tests on the three object detection models mentioned above using three separate test datasets: RADDet, CARRADA and CRUW. These models were individually trained with four distinct datasets: the RADDet training set, the CARRADA training set, the CRUW training set, or the \textit{RadSimReal} dataset. 
Performance was assessed using Average Precision (AP) for class 'car' at IOU thresholds 0.1, 0.3 and 0.5. The AP was determined by the area under the precision-recall curve.
Table \ref{tab:ap_results_main} presents AP results, comparing performance between training and testing on individual datasets versus training with \textit{RadSimReal} and testing on different datasets. 
It additionally includes cross-dataset evaluation between RADDet and CARRADA, both characterized by a stationary radar setup (unlike CRUW), with the primary difference lying in their scenes.

The results reveal several important insights. 
All three models trained using \textit{RadSimReal} exhibit performance on both the RADDet and CRUW test sets that closely resembles their performance when trained on the corresponding RADDet or CRUW training sets.
In evaluations with the CARRADA test set, models trained with \textit{RadSimReal} consistently outperform those trained with the CARRADA training set, likely due to the small size of the latter. 
Notably, models trained with RADDet experience a significant performance decline on the CARRADA test set compared to their performance when trained with \textit{RadSimReal}. These results demonstrate that object detection DNNs trained with \textit{RadSimReal} perform comparable to those trained on real data and even outperform DNNs trained on real data when subjected to cross-dataset evaluation or when dealing with limited training data.

Subsequently, we assess the performance of object detection models on the RADDet test set when trained using a combination of data from \textit{RadSimReal} and the training set of RADDet. The outcomes of this evaluation are presented in Table \ref{tab:ap_mix_real_and_synthetic}. The findings reveal that augmenting the \textit{RadSimReal} dataset with real datasets from RADDet does not yield a significant performance enhancement. This suggests that the domain shift from \textit{RadSimReal} data to real data is insignificant. 

Next, we provide qualitative examples  comparing between an object detection  model trained with \textit{RadSimReal} and one trained with real data. In Fig.~\ref{fig:qual_ex_det_score}, we showcase the detection scores at the output of the `U-Net' model for two examples taken from the RADDet test set, comparing the performance of the model trained with \textit{RadSimReal} against the same model trained with the RADDet training set. Each example is displayed in a separate row, featuring the original image alongside the output detection scores and bounding boxes of both models. Notably, the detection scores and boxes of both models resemble each other, indicating that the model trained with synthetic data delivers similar performance to the one trained exclusively on real data.

\begin{table}[]
\centering
\caption{Object detection AP at various IOU for `U-Net'  trained with combined \textit{RadSimReal}'s data (Sim) and real data (Real)}
\begin{tabular}{ccccc}
\toprule
Test Set & Train Set &  @0.1 & @0.3 & @0.5 \\
\midrule
 \midrule
\multirow{2}{*}{RADDet}  & Sim & 85.63 & 82.16 & 57.64 \\
 & Sim + Real & 86.09 & 83.38 & 58.13 \\
 \midrule
\multirow{2}{*}{CARRADA}  & Sim & 70.77 & 62.47 & 43.96\\
 & Sim + Real & 71.11 & 62.63 & 44.52 \\
\bottomrule
\end{tabular}
\vspace{-10pt}
\label{tab:ap_mix_real_and_synthetic}
\end{table}

The analysis in this section shows \textit{RadSimReal}'s success in bridging the object detection performance gap between synthetic and real data. It is important to note that the images generated by \textit{RadSimReal} are similar to those produced by other existing physical radar simulations. Consequently, training object detection models with other physical radar simulations could achieve a similar performance as with \textit{RadSimReal}. The significance of our study lies in unveiling this key discovery for the first time and introducing \textit{RadSimReal}, which holds advantages over existing simulations. It eliminates the need for in-depth knowledge of the radar design, typically undisclosed, and has faster run-time.
Additionally, we acknowledge that the synthetic to real performance gap can also be closed by generating data with generative methods such as GAN. However, these approaches have a major drawback compared to the physical simulation model; they require the collection of a large amount of real data for each distinct variation in radar type, its mounting configuration and environmental conditions.
\section{Conclusion}\label{sec:conclusion}
This paper introduces \textit{RadSimReal}, a novel physical radar simulation that generates synthetic radar images for training object detection DNNs. We have shown that the \textit{RadSimReal} images closely resemble real radar images both qualitatively and statistically. Most importantly, our results reveal that training object detection DNNs with these synthetic images and testing them on real data yield results similar to those obtained when training exclusively with real data.
Moreover, it attains superior performance in cross-dataset evaluations with different real datasets.

\textit{RadSimReal} offers distinct advantages over alternative methods of synthetic data generation. It can efficiently simulate diverse radar types  without the need for extensive real data collection, a process demanding substantial resources, or in-depth knowledge of proprietary radar implementation details, which are often confidential. Instead, it only requires a  measurement of the radar's PSF.
This work highlights the great potential of radar simulation in radar-based computer vision applications, paving the way for its increased adoption and further exploration in this field.
\newpage
{
    \small
    \bibliographystyle{ieeenat_fullname}
    \bibliography{main}

\begin{thebibliography}{47}
\providecommand{\natexlab}[1]{#1}
\providecommand{\url}[1]{\texttt{#1}}
\expandafter\ifx\csname urlstyle\endcsname\relax
  \providecommand{\doi}[1]{doi: #1}\else
  \providecommand{\doi}{doi: \begingroup \urlstyle{rm}\Url}\fi

\bibitem[Arnold et~al.(2022)Arnold, Bauhofer, Mandelli, Henninger, Schaich, Wild, and ten Brink]{arnold2022maxray}
Maximilian Arnold, M Bauhofer, Silvio Mandelli, Marcus Henninger, Frank Schaich, Thorsten Wild, and Stephan ten Brink.
\newblock Maxray: A raytracing-based integrated sensing and communication framework.
\newblock In \emph{2022 2nd IEEE International Symposium on Joint Communications \& Sensing (JC\&S)}, pages 1--7. IEEE, 2022.

\bibitem[Barnes et~al.(2020)Barnes, Gadd, Murcutt, Newman, and Posner]{barnes2020oxford}
Dan Barnes, Matthew Gadd, Paul Murcutt, Paul Newman, and Ingmar Posner.
\newblock The oxford radar robotcar dataset: A radar extension to the oxford robotcar dataset.
\newblock In \emph{2020 IEEE International Conference on Robotics and Automation (ICRA)}, pages 6433--6438. IEEE, 2020.

\bibitem[Bassem and Atef(2004)]{bassem2004matlab}
R~Mahafza Bassem and Z~Elsherbeni Atef.
\newblock Matlab simulations for radar systems design.
\newblock \emph{CRC, London}, 2004.

\bibitem[Bilik et~al.(2016)Bilik, Bialer, Villeval, Sharifi, Kona, Pan, Persechini, Musni, and Geary]{bilik2016automotive}
Igal Bilik, Oded Bialer, Shahar Villeval, Hasan Sharifi, Keerti Kona, Marcus Pan, Dave Persechini, Marcel Musni, and Kevin Geary.
\newblock Automotive mimo radar for urban environments.
\newblock In \emph{2016 IEEE Radar Conference (RadarConf)}, pages 1--6. IEEE, 2016.

\bibitem[Bochkovskiy et~al.(2020)Bochkovskiy, Wang, and Liao]{bochkovskiy2020yolov4}
Alexey Bochkovskiy, Chien-Yao Wang, and Hong-Yuan~Mark Liao.
\newblock Yolov4: Optimal speed and accuracy of object detection.
\newblock \emph{arXiv preprint arXiv:2004.10934}, 2020.

\bibitem[Buyuksalih et~al.(2017)Buyuksalih, Bayburt, Buyuksalih, Baskaraca, Karim, and Rahman]{buyuksalih20173d}
Ismail Buyuksalih, Serdar Bayburt, Gurcan Buyuksalih, AP Baskaraca, Hairi Karim, and Alias~Abdul Rahman.
\newblock 3d modelling and visualization based on the unity game engine--advantages and challenges.
\newblock \emph{ISPRS Annals of the Photogrammetry, Remote Sensing and Spatial Information Sciences}, 4:\penalty0 161--166, 2017.

\bibitem[Cook(2012)]{cook2012radar}
Charles Cook.
\newblock \emph{Radar signals: An introduction to theory and application}.
\newblock Elsevier, 2012.

\bibitem[Danzer et~al.(2019)Danzer, Griebel, Bach, and Dietmayer]{danzer20192d}
Andreas Danzer, Thomas Griebel, Martin Bach, and Klaus Dietmayer.
\newblock 2d car detection in radar data with pointnets.
\newblock In \emph{2019 IEEE Intelligent Transportation Systems Conference (ITSC)}, pages 61--66. IEEE, 2019.

\bibitem[de~Oliveira and Bekooij(2020)]{de2020generating}
Marcio L~Lima de Oliveira and Marco~JG Bekooij.
\newblock Generating synthetic short-range fmcw range-doppler maps using generative adversarial networks and deep convolutional autoencoders.
\newblock In \emph{2020 IEEE Radar Conference (RadarConf20)}, pages 1--6. IEEE, 2020.

\bibitem[Dong et~al.(2020)Dong, Wang, Zhang, and Liu]{dong2020probabilistic}
Xu Dong, Pengluo Wang, Pengyue Zhang, and Langechuan Liu.
\newblock Probabilistic oriented object detection in automotive radar.
\newblock In \emph{Proceedings of the IEEE/CVF Conference on Computer Vision and Pattern Recognition Workshops}, pages 102--103, 2020.

\bibitem[Dosovitskiy et~al.(2017)Dosovitskiy, Ros, Codevilla, Lopez, and Koltun]{dosovitskiy2017carla}
Alexey Dosovitskiy, German Ros, Felipe Codevilla, Antonio Lopez, and Vladlen Koltun.
\newblock Carla: An open urban driving simulator.
\newblock In \emph{Conference on robot learning}, pages 1--16. PMLR, 2017.

\bibitem[Dreher et~al.(2020)Dreher, Er{\c{c}}elik, B{\"a}nziger, and Knol]{dreher2020radar}
Maria Dreher, Eme{\c{c}} Er{\c{c}}elik, Timo B{\"a}nziger, and Alois Knol.
\newblock Radar-based 2d car detection using deep neural networks.
\newblock In \emph{2020 IEEE 23rd International Conference on Intelligent Transportation Systems (ITSC)}, pages 1--8. IEEE, 2020.

\bibitem[Dudek et~al.(2010)Dudek, Wahl, Kissinger, Weigel, and Fischer]{dudek2010millimeter}
Manuel Dudek, Ren{\'e} Wahl, Dietmar Kissinger, Robert Weigel, and Georg Fischer.
\newblock Millimeter wave fmcw radar system simulations including a 3d ray tracing channel simulator.
\newblock In \emph{2010 Asia-Pacific Microwave Conference}, pages 1665--1668. IEEE, 2010.

\bibitem[Feng et~al.(2019)Feng, Zhang, Kunert, and Wiesbeck]{feng2019point}
Zhaofei Feng, Shuo Zhang, Martin Kunert, and Werner Wiesbeck.
\newblock Point cloud segmentation with a high-resolution automotive radar.
\newblock In \emph{AmE 2019-Automotive meets Electronics; 10th GMM-Symposium}, pages 1--5. VDE, 2019.

\bibitem[Fidelis et~al.(2023)Fidelis, Reway, Ribeiro, Campos, Huber, Icking, Faria, and Sch{\"o}n]{fidelis2023generation}
Eduardo~C Fidelis, Fabio Reway, Herick Ribeiro, Pietro~L Campos, Werner Huber, Christian Icking, Lester~A Faria, and Torsten Sch{\"o}n.
\newblock Generation of realistic synthetic raw radar data for automated driving applications using generative adversarial networks.
\newblock \emph{arXiv preprint arXiv:2308.02632}, 2023.

\bibitem[Haitman and Bialer(2024)]{haitman2024boostrad}
Yuval Haitman and Oded Bialer.
\newblock Boostrad: Enhancing object detection by boosting radar reflections.
\newblock In \emph{Proceedings of the IEEE/CVF Winter Conference on Applications of Computer Vision}, pages 1638--1647, 2024.

\bibitem[Hess(2013)]{hess2013blender}
Roland Hess.
\newblock \emph{Blender foundations: The essential guide to learning blender 2.5}.
\newblock Taylor \& Francis, 2013.

\bibitem[Heusel et~al.(2017)Heusel, Ramsauer, Unterthiner, Nessler, and Hochreiter]{heusel2017gans}
Martin Heusel, Hubert Ramsauer, Thomas Unterthiner, Bernhard Nessler, and Sepp Hochreiter.
\newblock Gans trained by a two time-scale update rule converge to a local nash equilibrium.
\newblock \emph{Advances in neural information processing systems}, 30, 2017.

\bibitem[Hirsenkorn et~al.(2017)Hirsenkorn, Subkowski, Hanke, Schaermann, Rauch, Rasshofer, and Biebl]{hirsenkorn2017ray}
Nils Hirsenkorn, Paul Subkowski, Timo Hanke, Alexander Schaermann, Andreas Rauch, Ralph Rasshofer, and Erwin Biebl.
\newblock A ray launching approach for modeling an fmcw radar system.
\newblock In \emph{2017 18th International Radar Symposium (IRS)}, pages 1--10. IEEE, 2017.

\bibitem[Holder et~al.(2019)Holder, Linnhoff, Rosenberger, and Winner]{holder2019fourier}
Martin Holder, Clemens Linnhoff, Philipp Rosenberger, and Hermann Winner.
\newblock The fourier tracing approach for modeling automotive radar sensors.
\newblock In \emph{2019 20th International Radar Symposium (IRS)}, pages 1--8. IEEE, 2019.

\bibitem[Instruments(2022{\natexlab{a}})]{TI_PROTOTYPE_1}
Texas Instruments.
\newblock {AWR1843} data sheet, product information and support.
\newblock Datasheet, 2022{\natexlab{a}}.

\bibitem[Instruments(2022{\natexlab{b}})]{TI_PROTOTYPE_2}
Texas Instruments.
\newblock {TI} mmwave-sdk software development kit ({SDK}).
\newblock Datasheet, 2022{\natexlab{b}}.

\bibitem[Kim et~al.(2020)Kim, Cho, Kim, Kim, and Lee]{kim2020yolo}
Woosuk Kim, Hyunwoong Cho, Jongseok Kim, Byungkwan Kim, and Seongwook Lee.
\newblock Yolo-based simultaneous target detection and classification in automotive fmcw radar systems.
\newblock \emph{Sensors}, 20\penalty0 (10):\penalty0 2897, 2020.

\bibitem[Knott et~al.(2004)Knott, Schaeffer, and Tulley]{knott2004radar}
Eugene~F Knott, John~F Schaeffer, and Michael~T Tulley.
\newblock \emph{Radar cross section}.
\newblock SciTech Publishing, 2004.

\bibitem[Kraus et~al.(2020)Kraus, Scheiner, Ritter, and Dietmayer]{kraus2020using}
Florian Kraus, Nicolas Scheiner, Werner Ritter, and Klaus Dietmayer.
\newblock Using machine learning to detect ghost images in automotive radar.
\newblock In \emph{2020 IEEE 23rd International Conference on Intelligent Transportation Systems (ITSC)}, pages 1--7. IEEE, 2020.

\bibitem[Meyer et~al.(2021)Meyer, Kuschk, and Tomforde]{meyer2021graph}
Michael Meyer, Georg Kuschk, and Sven Tomforde.
\newblock Graph convolutional networks for 3d object detection on radar data.
\newblock In \emph{Proceedings of the IEEE/CVF International Conference on Computer Vision}, pages 3060--3069, 2021.

\bibitem[Ngo et~al.(2021)Ngo, Bauer, and Resch]{ngo2021multi}
Anthony Ngo, Max~Paul Bauer, and Michael Resch.
\newblock A multi-layered approach for measuring the simulation-to-reality gap of radar perception for autonomous driving.
\newblock In \emph{2021 IEEE International Intelligent Transportation Systems Conference (ITSC)}, pages 4008--4014. IEEE, 2021.

\bibitem[Ouaknine et~al.(2021)Ouaknine, Newson, Rebut, Tupin, and P{\'e}rez]{ouaknine2021carrada}
Arthur Ouaknine, Alasdair Newson, Julien Rebut, Florence Tupin, and Patrick P{\'e}rez.
\newblock Carrada dataset: Camera and automotive radar with range-angle-doppler annotations.
\newblock In \emph{2020 25th International Conference on Pattern Recognition (ICPR)}, pages 5068--5075. IEEE, 2021.

\bibitem[Paek et~al.(2022)Paek, Kong, and Wijaya]{paek2022k}
Dong-Hee Paek, Seung-Hyun Kong, and Kevin~Tirta Wijaya.
\newblock K-radar: 4d radar object detection dataset and benchmark for autonomous driving in various weather conditions.
\newblock \emph{arXiv preprint arXiv:2206.08171}, 2022.

\bibitem[Rebut et~al.(2022)Rebut, Ouaknine, Malik, and P{\'e}rez]{rebut2022raw}
Julien Rebut, Arthur Ouaknine, Waqas Malik, and Patrick P{\'e}rez.
\newblock Raw high-definition radar for multi-task learning.
\newblock In \emph{Proceedings of the IEEE/CVF Conference on Computer Vision and Pattern Recognition}, pages 17021--17030, 2022.

\bibitem[Redmon et~al.(2016)Redmon, Divvala, Girshick, and Farhadi]{redmon2016you}
Joseph Redmon, Santosh Divvala, Ross Girshick, and Ali Farhadi.
\newblock You only look once: Unified, real-time object detection.
\newblock In \emph{Proceedings of the IEEE conference on computer vision and pattern recognition}, pages 779--788, 2016.

\bibitem[Rohling(1983)]{rohling1983radar}
Hermann Rohling.
\newblock Radar cfar thresholding in clutter and multiple target situations.
\newblock \emph{IEEE transactions on aerospace and electronic systems}, pages 608--621, 1983.

\bibitem[Scheiner et~al.(2019)Scheiner, Appenrodt, Dickmann, and Sick]{scheiner2019radar}
Nicolas Scheiner, Nils Appenrodt, J{\"u}rgen Dickmann, and Bernhard Sick.
\newblock Radar-based road user classification and novelty detection with recurrent neural network ensembles.
\newblock In \emph{2019 IEEE Intelligent Vehicles Symposium (IV)}, pages 722--729. IEEE, 2019.

\bibitem[Sch{\"o}ffmann et~al.(2021)Sch{\"o}ffmann, Ubezio, B{\"o}hm, M{\"u}hlbacher-Karrer, and Zangl]{schoffmann2021virtual}
Christian Sch{\"o}ffmann, Barnaba Ubezio, Christoph B{\"o}hm, Stephan M{\"u}hlbacher-Karrer, and Hubert Zangl.
\newblock Virtual radar: Real-time millimeter-wave radar sensor simulation for perception-driven robotics.
\newblock \emph{IEEE Robotics and Automation Letters}, 6\penalty0 (3):\penalty0 4704--4711, 2021.

\bibitem[Schumann et~al.(2018)Schumann, Hahn, Dickmann, and W{\"o}hler]{schumann2018semantic}
Ole Schumann, Markus Hahn, J{\"u}rgen Dickmann, and Christian W{\"o}hler.
\newblock Semantic segmentation on radar point clouds.
\newblock In \emph{2018 21st International Conference on Information Fusion (FUSION)}, pages 2179--2186. IEEE, 2018.

\bibitem[Sch{\"u}{\ss}ler et~al.(2021)Sch{\"u}{\ss}ler, Hoffmann, Br{\"a}unig, Ullmann, Ebelt, and Vossiek]{schussler2021realistic}
Christian Sch{\"u}{\ss}ler, Marcel Hoffmann, Johanna Br{\"a}unig, Ingrid Ullmann, Randolf Ebelt, and Martin Vossiek.
\newblock A realistic radar ray tracing simulator for large mimo-arrays in automotive environments.
\newblock \emph{IEEE Journal of Microwaves}, 1\penalty0 (4):\penalty0 962--974, 2021.

\bibitem[Sheeny et~al.(2021)Sheeny, De~Pellegrin, Mukherjee, Ahrabian, Wang, and Wallace]{sheeny2021radiate}
Marcel Sheeny, Emanuele De~Pellegrin, Saptarshi Mukherjee, Alireza Ahrabian, Sen Wang, and Andrew Wallace.
\newblock Radiate: A radar dataset for automotive perception in bad weather.
\newblock In \emph{2021 IEEE International Conference on Robotics and Automation (ICRA)}, pages 1--7. IEEE, 2021.

\bibitem[Skolnik(1980)]{skolnik1980introduction}
Merrill~Ivan Skolnik.
\newblock Introduction to radar systems.
\newblock \emph{New York}, 1980.

\bibitem[Thieling et~al.(2020)Thieling, Frese, and Ro{\ss}mann]{thieling2020scalable}
J{\"o}rn Thieling, Susanne Frese, and J{\"u}rgen Ro{\ss}mann.
\newblock Scalable and physical radar sensor simulation for interacting digital twins.
\newblock \emph{IEEE Sensors Journal}, 21\penalty0 (3):\penalty0 3184--3192, 2020.

\bibitem[Turin(1960)]{turin1960introduction}
George Turin.
\newblock An introduction to matched filters.
\newblock \emph{IRE transactions on Information theory}, 6\penalty0 (3):\penalty0 311--329, 1960.

\bibitem[Wang et~al.(2020)Wang, Goldluecke, and Anklam]{wang2020l2r}
Leichen Wang, Bastian Goldluecke, and Carsten Anklam.
\newblock L2r gan: Lidar-to-radar translation.
\newblock In \emph{Proceedings of the Asian Conference on Computer Vision}, 2020.

\bibitem[Wang et~al.(2021{\natexlab{a}})Wang, Jiang, Li, Hwang, Xing, and Liu]{wang2021rodnet}
Yizhou Wang, Zhongyu Jiang, Yudong Li, Jenq-Neng Hwang, Guanbin Xing, and Hui Liu.
\newblock Rodnet: A real-time radar object detection network cross-supervised by camera-radar fused object 3d localization.
\newblock \emph{IEEE Journal of Selected Topics in Signal Processing}, 15\penalty0 (4):\penalty0 954--967, 2021{\natexlab{a}}.

\bibitem[Wang et~al.(2021{\natexlab{b}})Wang, Jiang, Li, Hwang, Xing, and Liu]{wang2021rodnet2}
Yizhou Wang, Zhongyu Jiang, Yudong Li, Jenq-Neng Hwang, Guanbin Xing, and Hui Liu.
\newblock Rodnet: A real-time radar object detection network cross-supervised by camera-radar fused object 3d localization.
\newblock \emph{IEEE Journal of Selected Topics in Signal Processing}, 15\penalty0 (4):\penalty0 954--967, 2021{\natexlab{b}}.

\bibitem[Weston et~al.(2021)Weston, Jones, and Posner]{weston2021there}
Rob Weston, Oiwi~Parker Jones, and Ingmar Posner.
\newblock There and back again: Learning to simulate radar data for real-world applications.
\newblock In \emph{2021 IEEE International Conference on Robotics and Automation (ICRA)}, pages 12809--12816. IEEE, 2021.

\bibitem[Wheeler et~al.(2017)Wheeler, Holder, Winner, and Kochenderfer]{wheeler2017deep}
Tim~A Wheeler, Martin Holder, Hermann Winner, and Mykel~J Kochenderfer.
\newblock Deep stochastic radar models.
\newblock In \emph{2017 IEEE Intelligent Vehicles Symposium (IV)}, pages 47--53. IEEE, 2017.

\bibitem[Zhang et~al.(2021)Zhang, Nowruzi, and Laganiere]{zhang2021raddet}
Ao Zhang, Farzan~Erlik Nowruzi, and Robert Laganiere.
\newblock Raddet: Range-azimuth-doppler based radar object detection for dynamic road users.
\newblock In \emph{2021 18th Conference on Robots and Vision (CRV)}, pages 95--102. IEEE, 2021.

\bibitem[Zhang et~al.(2020)Zhang, Li, and Wenger]{zhang2020object}
Guoqiang Zhang, Haopeng Li, and Fabian Wenger.
\newblock Object detection and 3d estimation via an fmcw radar using a fully convolutional network.
\newblock In \emph{ICASSP 2020-2020 IEEE International Conference on Acoustics, Speech and Signal Processing (ICASSP)}, pages 4487--4491. IEEE, 2020.

\end{thebibliography}
}
\newpage
\appendix
\maketitlesupplementary
\section{PSF and Noise  Variance Measurement} 
This section explains the process of measuring the radar's Point Spread Function (PSF) and noise variance, essential components for \textit{RadSimReal} as elaborated in Section \ref{sec:sim_metthod}.
To obtain the PSF, a radar measurement can be conducted in a scenario featuring a narrow and stationary object, like a pole, positioned in an isolated area where there are no prominent reflecting objects nearby. The pole, characterized by narrow spread in distance, azimuth angle, and Doppler frequency, can be treated as an approximation of a point reflector. If possible, using a radar corner reflector as the isolated target is preferred. A radar corner reflector is specifically designed to exhibit an exceptionally narrow spread in all dimensions \cite{knott2004radar}.

To measure the PSF, multiple radar tensors that capture the same scenario as described above are collected at various time instances when both the radar and the observed object remain stationary. These tensors are then averaged to reduce the noise in the PSF measurement. Subsequently, the truncated PSF utilized in \textit{RadSimReal} is derived by extracting a 3D segment from the averaged radar tensor, centered around the reflection point of the narrow object. The intensity of the PSF diminishes rapidly from its central point.
Each dimension of the PSF is truncated at a point where its intensity significantly falls below the radar's noise variance (the noise variance in the tensor without averaging).

\begin{figure*}
  \centering
   \includegraphics[width=1.0\linewidth]{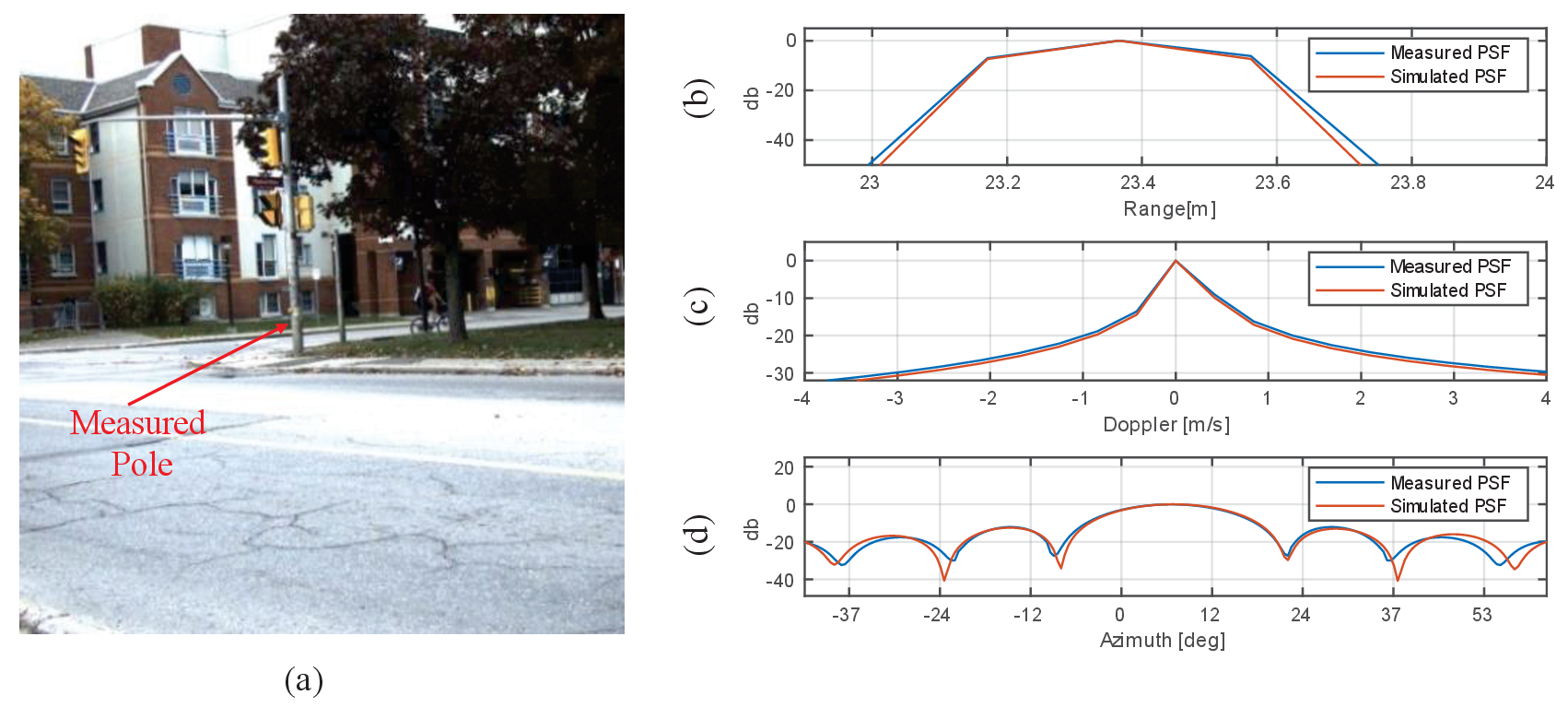}
   \vspace{-20pt}
   \caption{Measurement of the radar's PSF using a pole. (a) Image depicting the scenario and the employed pole. PSF slices in range, Doppler, and azimuth angle of the measured PSF compared with the PSF obtained through conventional radar simulation in (b), (c), and (d), respectively.}
   \label{fig:psf_measure}
\end{figure*}

Fig.~\ref{fig:psf_measure} provides a demonstration of the PSF measurement for the radar utilized in the RADDet dataset.  Fig.~\ref{fig:psf_measure}(a) presents a camera image of a scene from RADDet featuring a pole that was used for the PSF measurement. Figs.~\ref{fig:psf_measure}(b), (c), and (d) display the measured PSF slices in range, Doppler, and azimuth angle, respectively, alongside corresponding slices of a PSF obtained through conventional simulation of the radar in the RADDet dataset (as detailed in Fig. \ref{fig:sim_block_diagram}(a)+(b)). The figure illustrates that the measured PSF closely resembles the simulated radar's PSF.

It is worth noting that the simulation method employed to derive the reference PSF in Figs.~\ref{fig:psf_measure} necessitates an in-depth understanding of the specific radar hardware design and processing algorithms. This information is not always disclosed by radar suppliers. In contrast, the measurement procedure outlined above enables the acquisition of the PSF through a straightforward measurement that does not require detailed knowledge of the radar design.

\begin{figure}
  \centering
   \includegraphics[width=1.0\linewidth]{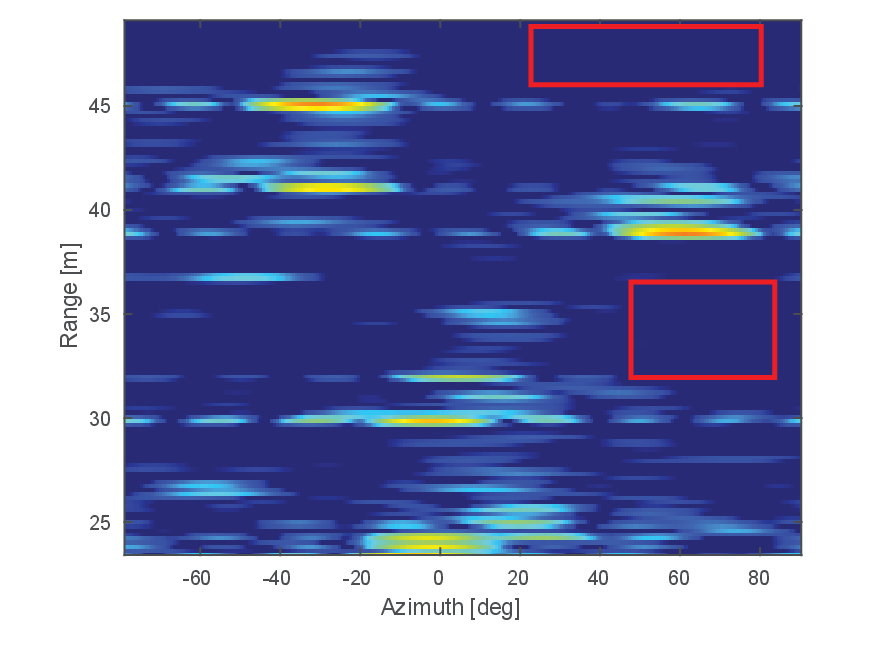}
   \vspace{-20pt}
   \caption{Measurement of radar noise variance from radar image in the RADDet dataset. Red rectangles in the radar image mark reflection-free regions utilized for noise variance measurement.}
   \label{fig:noise_measure}
\end{figure}

Next, we proceed to elucidate how to measure the radar's noise variance, a prerequisite for \textit{RadSimReal} as detailed in Section \ref{sec:sim_metthod}. The noise variance can be determined by identifying a region within the radar tensor that lacks any objects. This specific portion of the radar tensor comprises only noise, allowing the calculation of noise variance by assessing the variance of the tensor cells within this region. Fig.~\ref{fig:noise_measure} illustrates an instance of a radar image from the RADDet dataset, with red rectangles indicating sections without reflections that can be utilized for measuring the noise variance.

\section{Equivalency Between Outputs of Conventional Simulation and \textit{RadSimReal}}\label{sec::sim_equivalence}
In Section \ref{sec:sim_metthod}, we claimed that the identical output tensor from the conventional simulation (Fig. \ref{fig:sim_block_diagram}(a)+(b)) could be achieved by convolving the radar's Point Spread Function (PSF) with the 3D reflection points in the scene (Fig. \ref{fig:sim_block_diagram}(a)+(c)). This section provides an explanation for the validity of this equivalence. We first provide an intuitive understanding of this equivalence through a straightforward example in Section \ref{sec:equivalence_intuition}. Subsequently, in Section \ref{sec:equivalence_mathematic}, we present a formal mathematical derivation of this equivalence.

\subsection{Intuitive Explanation of Simulations  Equivalence}\label{sec:equivalence_intuition}
\begin{figure}
  \centering
   \includegraphics[width=1.0\linewidth]{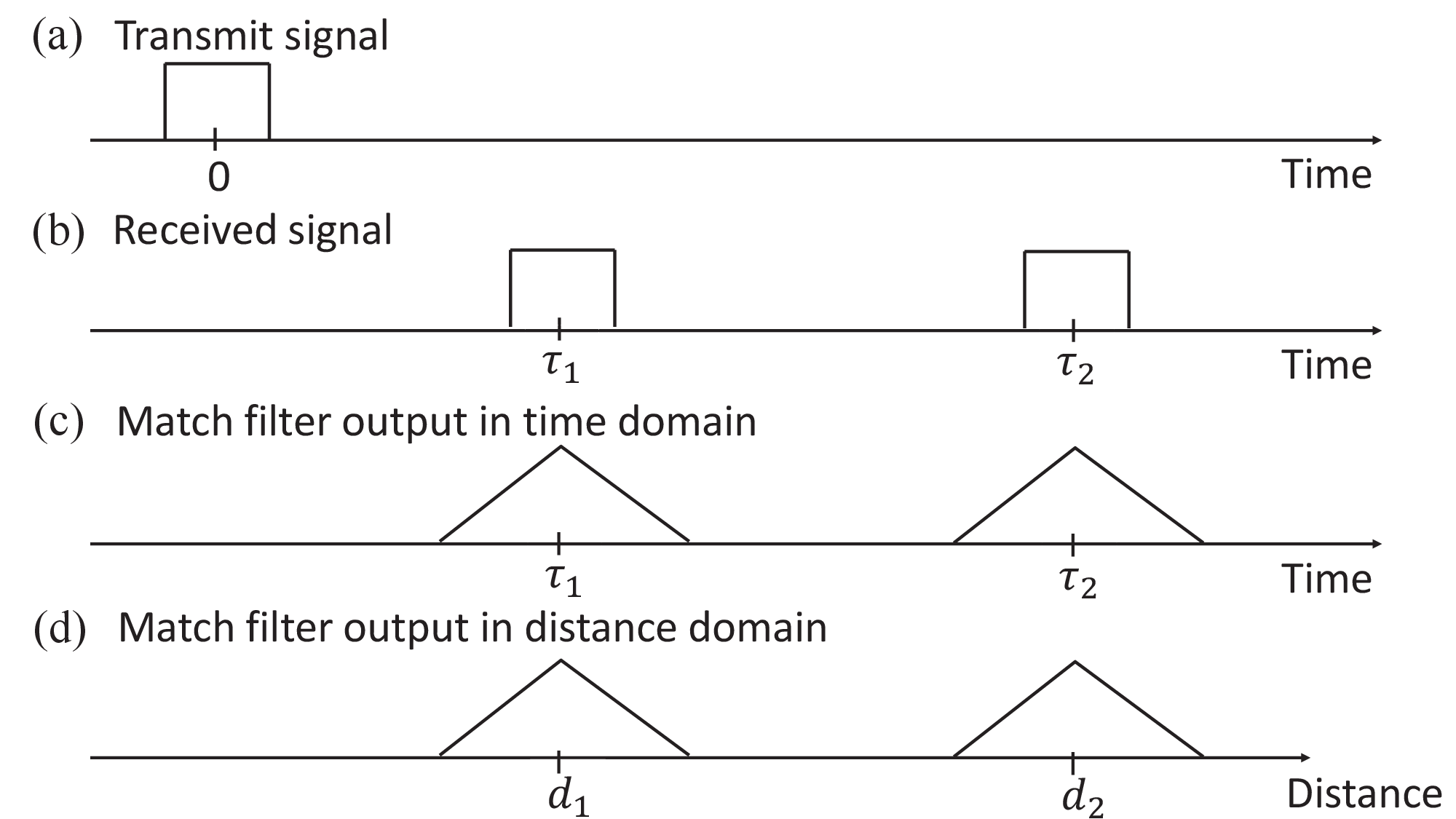}
   \vspace{-15pt}
   \caption{Illustration of the conventional simulation processes with a simplified Example. (a) Transmitted pulse. (b) Received signal from two reflections at delays $\tau_1,\tau_2$. (c) Output of receiver processing obtained by correlating the received signal with the transmitted signal (match filter). (d) Match filter output converted from time to distance scale.}
   \label{fig:conv_sim_illust}
\end{figure}

We explain the equivalence between the conventional simulation and \textit{RadSimReal} through a simplified example of a radar signal. The conventional radar simulation of this example is depicted in Fig.~\ref{fig:conv_sim_illust}. Fig.~\ref{fig:conv_sim_illust}(a) illustrates a transmitted radar pulse signal at time zero. In Fig.~\ref{fig:conv_sim_illust}(b), the received signal is illustrated for a scenario involving two reflection points. It is evident that the received signal is essentially the transmitted signal but delayed by $\tau_1$ and $\tau_2$. These delays represent the times taken for the signal to travel from the radar to the first and second reflection points and back. Importantly, these time delays  are proportional to the distances of the first and second reflection points denoted by $d_1$ and $d_2$, respectively. To obtain the received energy for each delay (distance) hypothesis, the radar employs a match filter on the received signal \cite{cook2012radar,turin1960introduction}. The match filter operation is a correlation between the received signal and the transmitted signal. Fig.~\ref{fig:conv_sim_illust}(c) displays the result of the match filter in this example.
The signal propagation times, $\tau_1$ and $\tau_2$, are proportional to the reflection points distances. Hence the time scale in Fig.~\ref{fig:conv_sim_illust}(c) can be converted to distance. The corresponding match filter output as a function of distance is depicted in Fig.~\ref{fig:conv_sim_illust}(d).

\begin{figure}
  \centering
   \includegraphics[width=1.0\linewidth]{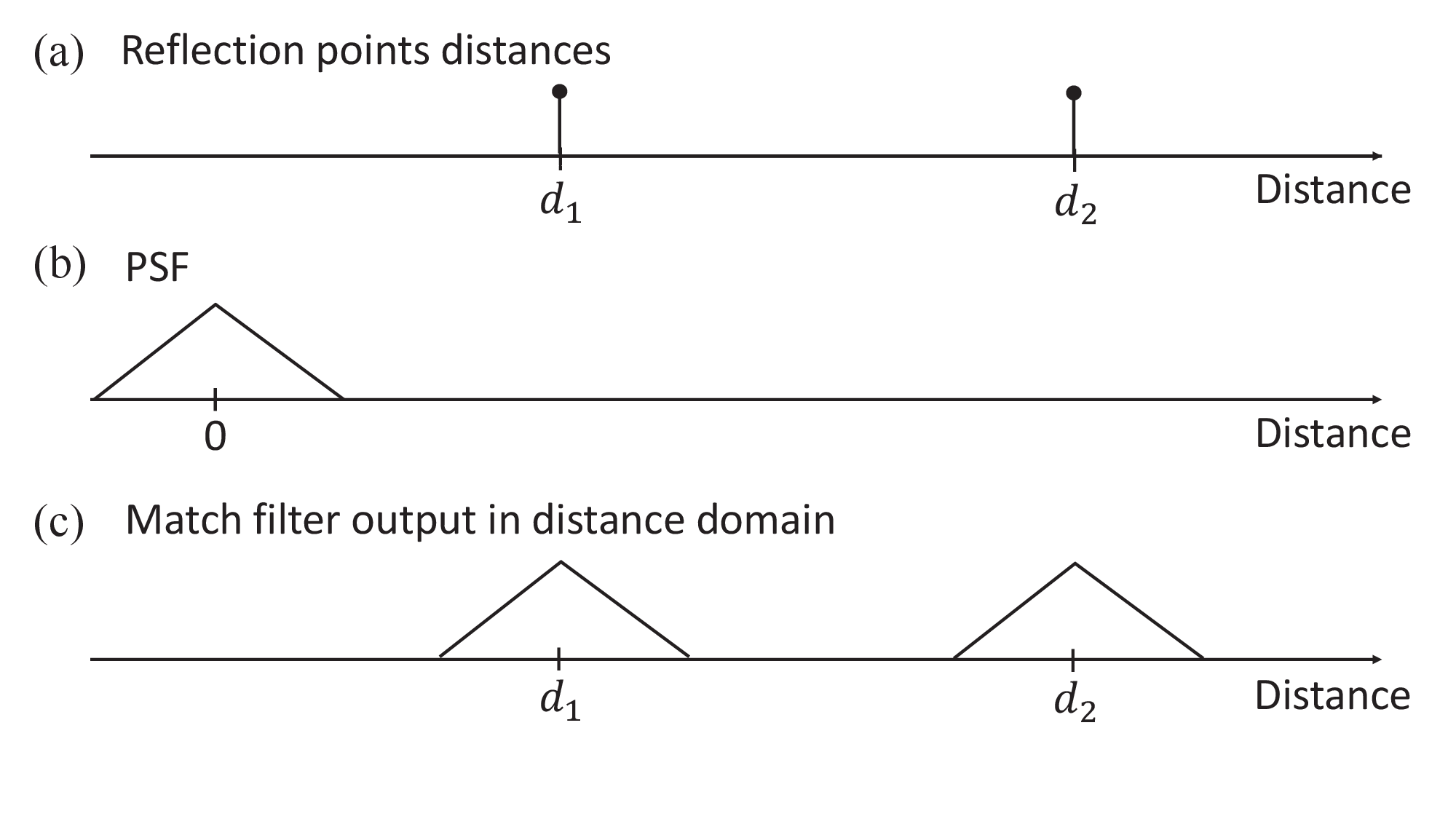}
   \vspace{-30pt}
   \caption{Illustration of the \textit{RadSimReal} processes achieving an identical output as the conventional simulation in the simplified example shown in Fig.~\ref{fig:conv_sim_illust}. (a) Representation of two reflection points using Kronecker delta functions centered at reflection points' distances $d_1$ and $d_2$. (b) The radar's PSF as a function of distance. (c) The result of convolution between the reflection points in (a) and the PSF in (b). The output in (c) is identical to the output of the conventional simulation shown in Fig.~\ref{fig:conv_sim_illust}(d).}
   \vspace{-10pt}
   \label{fig:our_sim_illust}
\end{figure}

Fig.~\ref{fig:our_sim_illust} illustrates the simulation conducted by \textit{RadSimReal} for the same example depicted in Fig.~\ref{fig:conv_sim_illust}. In Fig.~\ref{fig:our_sim_illust}(a), Kronecker delta functions at the distances $d_1$ and $d_2$ of the two reflection points are presented. Fig.~\ref{fig:our_sim_illust}(b) illustrates the radar's PSF, which is the auto-correlation of the transmitted signal. Fig.~\ref{fig:our_sim_illust}(c) displays the output of the convolution between the delta functions in Fig.~\ref{fig:our_sim_illust}(a) and the PSF in Fig.~\ref{fig:our_sim_illust}(b). It is evident that the match filter output in Fig.~\ref{fig:conv_sim_illust}(d) is identical to the result in Fig.~\ref{fig:our_sim_illust}(c). Thus, the output of the conventional simulation can be achieved through convolution between the radar's PSF and the reflection points, represented as delta functions at the distances of the reflection points. This approach is the simulation methodology employed by \textit{RadSimReal}.

While the representation in Fig.~\ref{fig:conv_sim_illust} and \ref{fig:our_sim_illust} show a one-dimensional match filter applied to a simplified single transmitted radar pulse, practical automotive radars involve extending the transmitted signal and match filter across a sequence of pulses and multiple antennas. Consequently, the result is a multidimensional output tensor with dimensions of range (distance), Doppler, and angle, rather than a single-dimensional output. This tensor is the one discussed in Section\ref{sec:sim_metthod}.
Nevertheless, the processes in each dimension can be separable. Therefore, the fundamental principle remains unchanged: the output radar tensor can be calculated by convolving a 3D PSF with dimensions for range, Doppler, and angle with delta functions in the 3D space having the same dimensions. The delta functions are positioned in this space at the reflection points' range, Doppler, and angle. A mathematical derivation of this equivalence is detailed in Section \ref{sec:equivalence_mathematic}.

\subsection{Mathematical Derivation of the Simulations Equivalence}\label{sec:equivalence_mathematic}
The radar emits a periodic sequence of short signals through multiple antennas, and this signal can be characterized in three dimensions as $s(n,m,q)$, where the indices $n$, $m$, and $q$ correspond to distinct time scales. These time scales represent the time samples of the short signal, the time samples of the short signal periods within the repetition sequence, and the signal duration along the antenna array, respectively. 

The transmitted signal reflects off objects in the environment and returns to the radar with delays in each of the time scales, which are proportional to the reflection position and speed. Let $\tau_d^i$, $\tau_f^i$, and $\tau_{\theta}^i$ denote the $i^{th}$ reflection point's delays in the short signal duration, the delay in the period between the short signals, and the delay between the antennas. These delays correspond to the reflection point's distance (range), Doppler frequency (radial velocity), and angle, respectively. The received signal is an aggregation of the received signals from individual reflection points, each with its corresponding reflection intensity. The received signal samples along each of the three time dimensions can be expressed as follows:
\begin{equation}\label{eq:rx_seg}
r(n,m,q)=\sum_{i}\alpha^i s(n-\tau_d^i,m-\tau_f^i,q-\tau_{\theta}^i),
\end{equation}
where $\alpha^i$ represents the intensity of the $i^{th}$ reflection point.

The radar tensor, which represents the received energy in each distance, Doppler, and angle, is obtained by applying a match filter to the received signal \cite{cook2012radar,turin1960introduction}. The match filter is a 3D correlation between the received signal and the transmitted signal $s(n,m,q)$, in all three delays dimensions, which are proportional to the distance, Doppler, and angle. This correlation is expressed by:
\begin{multline}\label{eq:match_filter}
y(n,m,q)=
\sum_{c,u,k}r(n-k,m-u,q-c)s(k,u,c)=\\
\sum_{i,c,u,k}\alpha_i
s(n-k-\tau_d^i,m-u-\tau_f^i,q-c-\tau_{\theta}^i)s(k,u,c).
\end{multline}
The radar's PSF is obtained by the auto-correlation of the transmitted signal given by 
\begin{equation}\label{eq:psf}
x(n,m,q)\triangleq\sum_{c,u,k}s(n-k,m-u,q-c)s(k,u,c).
\end{equation}
Then by substituting \eqref{eq:psf} into \eqref{eq:match_filter} we obtained that the match filter output can be expressed by 
\begin{equation}\label{eq:mf_psf}
y(n,m,q)=\sum_{i}\alpha_i x(n-\tau_d^i,m-\tau_f^i,q-\tau_{\theta}^i).
\end{equation}

From \eqref{eq:mf_psf}, it becomes evident that the radar tensor is a superposition of the radar's PSF shifted by the delays' of the reflection points and scaled by their intensities. This relationship can equivalently be expressed as:
\begin{equation}\label{eq:mf_conv}
y(n,m,q)=x(n,m,q) * \sum_{i}\alpha_i\delta(n-\tau_d^i,m-\tau_f^i,q-\tau_{\theta}^i),
\end{equation}
where the symbol $*$ denotes a convolution operation, and $\delta(n-\tau_d^i,m-\tau_f^i,q-\tau_{\theta}^i)$ is a 3D Kronecker delta function. This function takes a value of 1 when $n=\tau_d^i$, $m=\tau_f^i$, and $q=\tau_{\theta}^i$, and is zero elsewhere. Therefore, rather than obtaining the radar tensor through match filtering, as shown in \eqref{eq:match_filter}, it can equivalently be obtained by convolving the radar's PSF, $x(n,m,q)$, with reflection points that are represented by 3D Kronecker delta functions that are shifted by the 3D delays' ($\tau_d^i,\tau_f^i,\tau_{\theta}^i$) of the reflection points and scaled by their intensities ($\alpha_i$), as expressed in \eqref{eq:mf_conv}. 
The 3D delays are directly proportional to the range, Doppler frequency, and angle of the reflections. Consequently, the dimensions of the radar tensor are eventually transformed from delays to range, Doppler, and angle. This equivalent approach is the methodology utilized in deriving the tensor in \textit{RadSimReal}.

\section{Simulation Computation Complexity}
\begin{table}[h]
\caption{Object detection average precision (AP) on RADDet with original train-test split vs. our split}\label{tab:raddet_org_split}
\centering
\vspace{-5pt}
\resizebox{\columnwidth}{!}{
\begin{tabular}{ccccccc}
\toprule
       & \multicolumn{3}{c}{AP Original Split} & \multicolumn{3}{c}{AP Our Split}  \\
\cmidrule(lr){2-4} \cmidrule(lr){5-7}
Method                        & @ 0.1       & @ 0.3       & @ 0.5      & @ 0.1              & @ 0.3             & @ 0.5\\
\midrule
RADDet   & 93.82 & 88.03 & 68.71 &83.69 & 72.96 & 47.95 \\
\midrule
Probabilistic   & 92.87 & 86.36 & 66.30 &  83.31  &  74.80 & 40.68\\
\midrule
U-Net  & 94.68 &  89.59 &   71.00 & 84.76 & 83.01 & 55.53 \\
\bottomrule
\end{tabular}
}
\end{table}
\label{sec:sim_compute_complexity}
In this section, we assess the computational complexity of conventional radar simulation in comparison to \textit{RadSimReal}. Both simulations initiate with the shared step of generating reflection points in the environment (as depicted in Fig. \ref{fig:sim_block_diagram}(a)). The runtime of this phase depends on the graphics simulation engine and can be very fast, even in real-time. The significant run-time difference lies in transforming reflection points into radar images (parts (b) and (c) of Fig. \ref{fig:sim_block_diagram}), which are evaluated next.

The initial phase of the conventional simulation involves the generation of received samples per radar image frame. These samples result from aggregating the received signals from individual reflection points, leading to a computational complexity of $O(N_{p}N_{r})$, where $N_{r}$ represents the number of received samples per radar image, and $N_{p}$ denotes the number of reflection points in the scenario. The number of received samples per frame, $N_r$, is directly proportional to the number of cells in the radar tensor, denoted as $N_s$. Consequently, the complexity of the first part of the conventional simulation can be expressed as $O(N_{p}N_{s})$.

In the subsequent stage of the conventional radar simulation, signal processing algorithms are applied to the received signal to generate the radar tensor. These algorithms coherently combine received signal samples for each range, angle, and Doppler cell in the radar tensor. This process, called match filtering, is efficiently executed through a series of Fast Fourier Transform (FFT) operations in range, Doppler, and angle, resulting in an overall complexity that is lower bounded by $O(N_{s}\log({N_{s}}))$.

The total complexity of the conventional simulation is derived by summing the complexities of the two aforementioned parts. This yields a complexity of $O(N_{s}(N_p+\log({N_{s}})))$. Since $N_p$ is significantly larger than $\log({N_s})$, the complexity of the conventional simulation approach simplifies to $O(N_{s}N_p)$.

Moving on, we proceed to compute the computational complexity of \textit{RadSimReal}, which involves performing convolution between the reflection points and the radar's PSF. The reflection points are sparsely distributed within the radar tensor, i.e., $N_s \gg N_p$. Consequently, the convolution between the reflection points and the PSF can be carried out as a sparse convolution. This operation entails aggregating the PSFs of reflection points, resulting in a complexity of $O(N_pN_{f})$, where $N_{f}$ denotes the number of cells in the radar's PSF.

Therefore, the complexity ratio between conventional simulations and \textit{RadSimReal} is expressed as $O((N_{s}N_p)/(N_pN_{f}))=O(N_s/N_{f})$, which represents the proportion of the entire radar tensor volume to the PSF volume. As detailed in Section \ref{sec:sim_metthod}, our simulation truncates the PSF to preserve $99\%$ of its energy, significantly reducing its volume. Consequently, this leads to a substantial reduction in complexity, exemplified by a ratio of 1250 for the radar utilized in the RADDet, CARRADA, and CRUW datasets.

We tested the average run time for generating radar images using  the conventional physical radar simulation and our simulation. These tests were conducted for the TI radar employed in the RADDet  \cite{TI_PROTOTYPE_1,TI_PROTOTYPE_2}. We implemented the simulations in Matlab 2020a, making use of the parallel processing toolbox. The computations were executed on a computer equipped with an Intel(R) Xeon(R) W-2235 CPU operating at 3.80GHz, alongside an Nvidia Quadro RTX 5000 GPU with 16GB of memory. The results reveal that the average run time for generating a radar images from reflection points with our simulation is 0.0105 second, and with the conventional physical simulation it takes 5.296 seconds. \textit{RadSimReal} generates images about 500 times faster than the conventional simulation, which is on the order of the factor 1250 that was obtained from the analysis above.
\begin{table}[t]
\caption{Instances in the RADDet dataset divided into distinct scenes.}
  \centering
  \begin{tabular}{cl}
    \toprule
    Scene ID & Frame Numbers  \\
    \midrule
0 & $0-439,559-724,1549-1971$\\
1 & $440-555,731-1548, 1972-2571$\\
2 & $2572-3038$ \\
3 & $3039-3437$\\
4 & $3438 -3653$\\
5 & $3654-4073$ \\
6 & $4074-4331$ \\
7 & $4332-5018, 5623 -6243$ \\
8 & $5019-5622,6244-6608$ \\
9 & $6609-8046$ \\
10 & $8047-8634$ \\
11 & $8635 - 9158$ \\
12 & $9159 -9437$ \\
13 & $9438-9745, 10175-10292$ \\
14 & $9746-10174$\\
    \bottomrule
  \end{tabular}
  \label{tab:RADDET_New_Split}
\end{table}

\section{RADDet Train-Test Set Split}
The performance evaluation conducted in Section \ref{sec:sim_validation} utilized a train-test set split of the RADDet dataset that differs from the split proposed in \cite{zhang2021raddet}. 
Table \ref{tab:raddet_org_split} displays the Average Precision (AP) results at IOU 0.1, 0.3, and 0.5 for the three object detection methods employed in the paper (`RADDet', `Probabilistic', and `U-Net'). The comparison is made between the original RADDet train-test split and our suggested split. The AP results are assessed on the test set within each respective split. The results reveal that all methods achieved significantly higher AP results with the original RADDet split compared to our split. This discrepancy in results can be attributed to the fact that the test and training images in the original split \cite{zhang2021raddet} were derived from the same scenarios, often with small temporal gaps. Consequently, a strong correlation is established between the test and training samples, leading to overfitting of all methods on the test set.

To address the issue of overfitting, we implemented a train-test set partitioning strategy that ensures distinct scenarios between the training and testing sets. The RADDet dataset comprises 15 unique scenes, each detailed in Table \ref{tab:RADDET_New_Split}. In our partitioning scheme, scenes 9 and 11 were designated for the test set, while the remaining scenes were used for the training set. The adjusted training set comprises a total of 17,021 cars compared to 16,755 in the original split. For the test set, we have 4,094 cars compared to 4,135 in the original split.

\end{document}